\documentclass[nonacm,authorversion]{acmart}
\usepackage{mathtools}
\usepackage{multirow}
\usepackage{subfigure}
\usepackage{wrapfig}
\usepackage{url}

\usepackage[noend,linesnumbered]{algorithm2e}
\usepackage[inline]{enumitem}
\usepackage{listings}
\usepackage{amsfonts}
\usepackage{pifont}% http://ctan.org/pkg/pifont
\newcommand{\cmark}{\ding{51}}%
\newcommand{\xmark}{\ding{55}}

\usepackage{lscape}
\usepackage{verbatim}
\AtBeginDocument{%
  }

\theoremstyle{acmdefinition}
\newtheorem{definition}{Definition}[section]

\newcommand{\supplementary}{{$^\text{\ref{supplementary}}$}} 

\newcommand\janik[1]{\textcolor{black}{#1}}
\newcommand\janikrev[1]{\textcolor{black}{#1}}
\newcommand\janikrr[1]{\textcolor{black}{#1}}

\newcommand\stefanierev[1]{\textcolor{black}{#1}}

\newcommand{\emphasizetext}[2][black]{%
  \begingroup
  \setlength{\fboxrule}{.6pt}%
  \setlength{\fboxsep}{1.5pt}%
  \fcolorbox{#1}{white}{\raisebox{0.5pt}{\textsf{\scriptsize\textbf{#2}}}}%
  \endgroup
}

\newcommand{\OB}[1]{%
  \emphasizetext{OB#1}%
}

\newcommand{\circled}[1]{%
  \textcircled{\scriptsize #1}%
}

\setcopyright{cc}
\setcctype{by}
\acmJournal{CSUR}
\acmYear{2025} \acmVolume{1} \acmNumber{1} \acmArticle{1}
\acmMonth{1} \acmPrice{}\acmDOI{10.1145/3743672}

\begin{document}

%\mainmatter  % start of an individual contribution

\title{A Survey on Event Prediction Methods from a Systems Perspective: Bringing Together Disparate Research Areas}

\author{Janik-Vasily Benzin}
\orcid{0000-0002-3979-400X}
\affiliation{%
  \institution{Technical University of Munich, Germany; TUM School of Computation, Information and Technology, Department of Informatics}
  \city{Garching}
  \country{Germany}
  %\postcode{85748}
}
\email{janik.benzin@tum.de}

\author{Stefanie Rinderle-Ma}
\orcid{0000-0001-5656-6108}
\affiliation{%
  \institution{Technical University of Munich, Germany; TUM School of Computation, Information and Technology, Department of Informatics}
  \city{Garching}
  \country{Germany}
  %\postcode{85748}
}
\email{stefanie.rinderle-ma@tum.de}

\renewcommand{\shortauthors}{JV. Benzin and S. Rinderle-Ma}


%The predicted events are presented to users 
%for evaluation with respect to 
%the desired future state. 
%Considering the diverse range of research areas and differing terminology, 
 \begin{abstract}

Event prediction is the ability of anticipating future events, i.e., future real-world occurrences, and aims to support the user in deciding on actions that change future events towards a desired state. An event prediction method learns the relation between features of past events and future events. It is applied to newly observed events to predict corresponding future events that are evaluated with respect to the user's desired future state. 
If the predicted future events do not comply with 
this state,
actions are taken towards achieving desirable future states. Evidently, event prediction is valuable in many application domains such as business and natural disasters. The diversity of application domains results in a diverse range of methods that are scattered across various research areas which, in turn, use different terminology for event prediction methods. 
Consequently, sharing methods and knowledge for developing future event prediction methods is restricted. To facilitate knowledge sharing on account of a comprehensive integration and assessment of event prediction methods, we take a systems perspective to integrate event prediction methods into a single system, elicit requirements, and assess existing work with respect to the requirements. Based on the assessment, we identify open challenges and discuss future research directions.

\end{abstract}


%\ccsdesc[500]{Applied computing~Business process management}

%\todo{gibt es auch Business Process Management als CCS concept?}

%\received{XXX}
%\received[revised]{XXX}
%\received[accepted]{2025-06-02}

\maketitle

% \nocite{*}

\section{Introduction}
\label{sec:intro}
\begin{figure}[htb!]
  \begin{minipage}{.64\linewidth}
    \centering
    \includegraphics[width=0.75\linewidth]{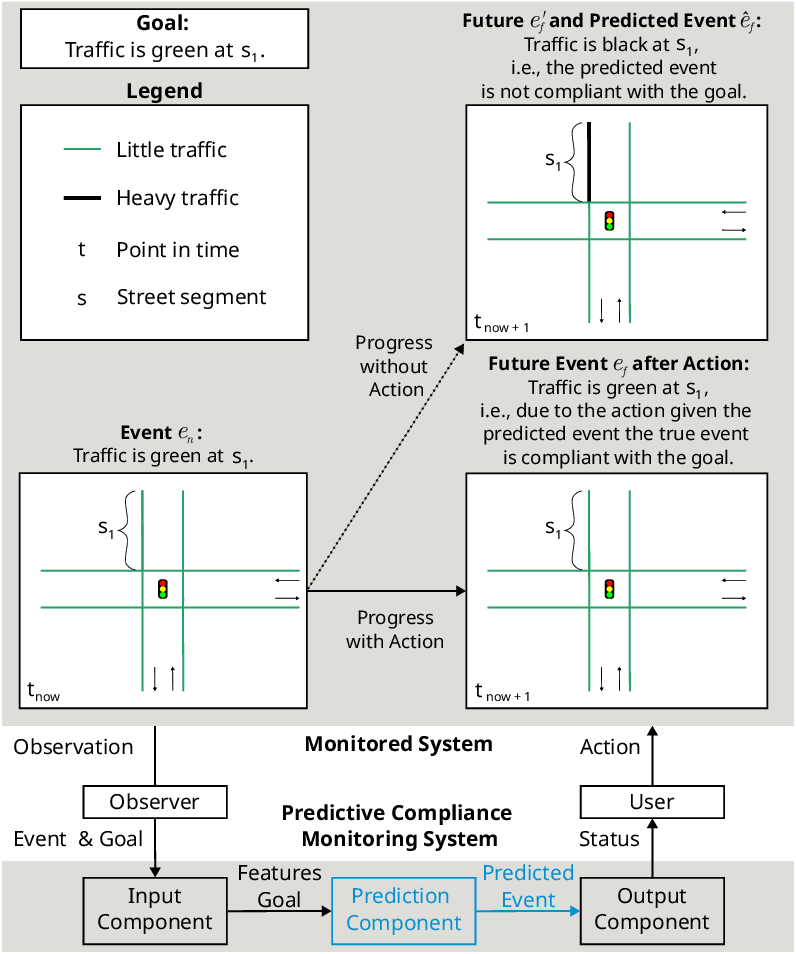}
    \caption
      {%
        EP as a component of a predictive compliance monitoring (PCM) system.%
        \label{fig:overview}%
        \Description{Illustration of event prediction as a component of a predictive compliance monitoring system. To begin with, we have a goal, e.g., that road segment $s_1$ has low traffic at an intersection. Hence, the intersection constitutes the monitored system. By observing changes in various factors of the monitored system (e.g., deteriorating traffic), the observer sends this change as an event and our goal to the input component of the predictive compliance monitoring system. The input component extracts features for the event prediction method inside the prediction component and the prediction components maps these to a set of predicted events. These predicted events are the input of the output component, which in turn represents these predicted events in a meaningful status to the user that can then decide on necessary actions. For example, if the predicted event shows that the traffic will deteriorate for the road segment at the intersection set in the goal meaning the predicted events do not comply with the goal, the user may decide to reroute traffic (the action) such that the monitored system maintains its low traffic rather than progressing towards the predicted heavy traffic.}%
      }%
  \end{minipage}\hfill
  \begin{minipage}{.34\linewidth}
    \centering
    \captionof{table}
      {\centering%
        Application domains of covered articles
        \label{tab:domain}%
      }
      \tiny 
    \begin{tabular}{lccc}
      \toprule
Domain & Short & Share & Count \\
\midrule
Business & B &0.42 & 109 \\
Engineering Systems & E &0.12 & 31 \\
Healthcare & H &0.09 & 23 \\
Political Events & P &0.09 & 23 \\
Cyber Systems & I &0.08 & 22 \\
Media & M &0.07 & 19 \\
Crime & C &0.06 & 15 \\
Natural Disasters & N &0.04 & 10 \\
Transportation & T &0.03 & 8 \\
\midrule
$ \sum $  & & 1.00 & 260 \\
\bottomrule
    \end{tabular}
    \vspace*{0.8cm}
    \captionof{table}
      {%
        List of Abbreviations
        \label{tab:abb}%
      }
    \tiny 
    \begin{tabular}{ll}
      \toprule
Abbreviation & Meaning \\
\midrule
EP & Event prediction \\
PPM & Predictive process monitoring \\
PCM & Predictive compliance monitoring \\
CM & Compliance monitoring \\

\bottomrule
    \end{tabular}
    
  \end{minipage}
\end{figure}

\janikrr{The rapid growth of data \cite{john_rydning_worldwide_2022} and major advances in enabling technologies such as cloud computing \cite{rashid2019cloud}, Internet of Things \cite{sadeeq2021iot}, and machine learning \cite{flouris2017issues} drives event prediction (EP) research (cf. Table \ref{tab:abb})}.
%is researched to a great extent. 
EP is the ability of anticipating events, i.e., future real-world occurrences.
To anticipate events, EP processes past events together with further relevant data for learning an EP method to map the processed input to predicted future events that are of interest to the respective application domain. Being able to anticipate events is highly beneficial to many application domains such as business, healthcare, transportation, crime and natural disasters \cite{zhao_event_2021}.
The value of knowing and ideally understanding predicted future events today lies in the ability of a user in the application domain to act on the predicted future events, e.g., by focusing police resources in areas with high crime risk. In order to unlock the potential of EP, several challenges have to be addressed \cite{zhao_event_2021}, including the knowledge on the true relationships between causes and effects of events \cite{zhao_event_2021}, heterogeneous multi-output predictions (e.g., predicting the time and location of future events) \cite{ramakrishnan_beating_2014}, complex dependencies among the predicted future events as they may interact with each other \cite{matsubara_fast_2012}, real-time stream of past events that requires continuous monitoring \cite{salfner_survey_2010,rinderle-ma_predictive_2022}, and 
%challenges in the past event data regarding data properties and 
data quality challenges \cite{rinderle-ma_predictive_2022}.

Existing work on EP has proposed a multitude of different solutions to those challenges. However, the solutions are often developed within one application domain, e.g., \janikrr{healthcare \cite{ronzani_unstructured_2022}}, and corresponding research area, \janikrr{e.g., predictive process monitoring (PPM).} 
The type of event and the laws and rules governing the occurrence of events greatly differ among the application domains, e.g., large-scale weather events governed by the laws and rules of meteorology in the domain of natural disasters versus small-scale business events of a business process governed by the respective business regulations and internal procedures in the domain of business \cite{allan_-line_1998}. Thus, EP research typically approaches the problem in an event-type-and domain-dependent manner \cite{zhao_event_2021,rinderle-ma_predictive_2022}.
Despite apparent advantages of this approach to EP, it hampers \janikrev{sharing} EP methods and solutions across
%the event prediction challenges with respect to 
 different application domains and corresponding research areas. \janikrr{Most importantly, the fragmentation in EP research significantly complicates any effort to draw a clear picture on where we are standing on our road towards applying EP methods to real-world problems and how we should proceed. In this paper, we aim at providing a clear picture on EP research, its fragmentation, current status, and remaining challenges by taking a \emph{systems perspective}. Taking a systems perspective results in a common architecture with requirements for EP methods that are generally valid across domains
 %despite different laws and rules. This enables 
%\janikrr{Let us start with an illustration on how the systems perspectives is beneficial 
and that enable assessing EP methods for real-world applications. }
%A comprehensive assessment of existing work sheds light on potential blind spots missed so far and establishes a well-founded understanding of the current research status for open challenges and future research directions.  We call the common architecture the \emph{predictive compliance monitoring} (PCM) system \cite{rinderle-ma_predictive_2022}. The PCM system serves as a blueprint to operationalize EP methods. The following example illustrates the PCM system
%To better understand EP through a system perspective, \autoref{fig:overview} depicts an intersection example from the transportation domain. The intersection has a road segment $s_1$ that is depicted at two points in time, the current point in time $t_{\text{now}} $ and a future point in time $t_{\text{now}+1}$. Our goal for the intersection with road segment $s_1$, the \emph{monitored system} in this example, is to maintain green traffic for $s_1$, i.e., on road segment $s_1$ little traffic should be maintained. Currently, event ``Traffic is green at $s_1$'' at $t_{\text{now}}$ \emph{complies with} our goal. This event together with the goal is observed by the \emph{observer}, e.g., a video camera directed at road segment $s_1$ and a user interface for setting goals. In general, the observer is an abstract interface between the monitored system in the real-world and the system for EP, i.e., it typically consists of a set of sensors, user interfaces and/or application programming interfaces. 

To illustrate the architecture at a high-level, \autoref{fig:overview} depicts an intersection example from the transportation domain. The intersection has a road segment $s_1$ \janikrr{that is the \emph{monitored system}}. $s_1$ is depicted at two points in time: the current point in time $t_{\text{now}} $ and \janikrr{two different future states at} a future point in time $t_{\text{now}+1}$. \janikrr{The first future state is labeled with both the ``future and predicted event'' to which $s_1$ will progress without an action. Hence, $s_1$ will and is predicted to face heavy traffic without an action. The second future state is labeled with ``future event with action'', Thus, $s_1$ will have little traffic after an action.}
\janikrr{Our \emph{event prediction goal}} (cf. goal in \autoref{fig:overview}) for the intersection with road segment $s_1$ is to maintain little traffic for $s_1$. Currently, event $e_n=$``Traffic is green at $s_1$'' at $t_{\text{now}}$ \emph{complies with} our goal. Event $e$ together with the goal is observed by the \emph{observer}, e.g., a video camera directed at road segment $s_1$, and a user interface for setting goals. In general, the observer is an abstract interface between the monitored system in the real-world and the system for EP. Therefore, the observer typically consists of a set of sensors, user interfaces and/or application programming interfaces. 

Then, the observer sends event $e_n$ and the goal to the \emph{input component} of the generalized \emph{predictive compliance monitoring} (PCM) system (\janik{cf. \autoref{tab:abb}}). \janikrr{The input component} processes the event and goal for subsequent prediction by the \emph{prediction component}. \emph{Predictive compliance monitoring} is concerned with continuously anticipating whether the monitored system complies to a goal or set of goals in the future and supporting the user in understanding and acting on the anticipated compliance status \cite{rinderle-ma_predictive_2022}. Thus, a generalized PCM system conceptualizes EP through its input component and prediction component, i.e., the EP method is conceptualized as the prediction component. After pre-processing and encoding the event in the input component, the prediction component correctly predicts $\hat{e}_f=$``Traffic is black at $s_1$'' at point in time $t_{\text{now}+1}$. Hence, the future progress of the monitored system is predicted to be non-compliant with our goal. 

The \emph{output component} of the generalized PCM system prepares a status for the user. \janikrr{Users are characterized by having the ability to take actions to steer the monitored system into a future direction compliant with the goal. For example,} an urban traffic control center is a user that decides on the necessary action to change the future towards a compliant status. For road segment $s_1$, the \janikrr{chosen action is to} reroute incoming traffic. Due to the action, the \janik{ground-true real event} (true event) at $t_{\text{now}+1}$ is \janikrr{$e_f=$}``Traffic is green at $s_1$''. Thus, the PCM system supports the user to maintain a future that is compliant with the goal. %The predicted $\hat{e}_f$``Traffic is black at $s_1$'' is non-compliant with the goal, but correctly predicts the future event $e_f^{\prime}$ that describes heavy traffic for $s_1$.  
To sum up, the user of the application domain is able to anticipate the future of the monitored system with respect to the goal and act accordingly on the account of the generalized PCM system. 

 \janikrr{Consider a practitioner that is faced with the problem of operationalizing the PCM system (cf. \autoref{fig:overview}). The practitioner is particularly interested in a quantification of how certain the method is in predicting the future event (\emph{uncertainty quantification}). There are at least eight existing EP methods that focus on the transportation domain (cf. Table \ref{tab:domain}). Yet, there are at least 252 other EP methods that may be better suited for operationalizing the actual PCM system with uncertainty quantification in spite of the different domain. Because the research landscape is fragmented, it is very hard to understand whether an EP method with a specific property may exist and whether it is applicable to the PCM system.  \cite{metzger_predictive_2017}, for example, propose an EP method based on neural networks that predicts wether a logistics process will meet a delivery deadline. Does this EP method fit the needs of the practitioner? Giving an answer is difficult for several reasons. First, \cite{metzger_predictive_2017} is missing in existing surveys on EP methods such as \cite{zhao_event_2021}. 
 %While \cite{zhao_event_2021} focuses on method classification and technical approaches to prediction, it does not unify the disparate research landscape through a systems perspective. 
 Second, \cite{metzger_predictive_2017} conceptualize the logistics process in a specific modeling notation that the practitioner has to know in order to analyze to what extent the method is applicable to the real-world problem. Third, the title contains the term \emph{business process}. Hence, if the practitioner is a civil engineer, the EP method proposed in \cite{metzger_predictive_2017} may quickly be discarded. Lastly, \cite{metzger_predictive_2017} do not tell that the deadline is predicted through a binary answer and that reliability estimates refers to uncertainty quantification. The former can be deduced from the implicit assumption pervading related methods from the same area: simple yes/no answers are sufficient to represent future events \cite{rinderle-ma_predictive_2022}.}  
 
\janikrr{To sum up, disparate research areas are difficult to navigate. Due to each area sharing its own conceptual models and implicit assumptions, the event prediction problem and the different solution methods seem incommensurable.} Different terminology for EP such as PPM (\janik{cf. Table \ref{tab:abb}}) in the field of business process management \cite{di_francescomarino_predictive_2018,marquez-chamorro_predictive_2018,teinemaa_outcome-oriented_2019} or anomaly prediction in the field of engineering systems \cite{lin_anomaly_2021} exacerbates the problem of cumbersome method and knowledge \janikrev{sharing} among various EP research areas. 
Here, taking a \textsl{systems perspective} on EP 
%leading to a system for event prediction and 
seems promising to place and integrate the diverse range of EP methods and, thus, foster knowledge and method sharing. 
The example in \autoref{fig:overview} indicates how the system perspective benefits the integration of EP methods by giving the method a general, yet practical context that covers important parts such as goals and actions. \janikrr{Moreover, it gives a clear definition of the components, functionalities, and interfaces necessary to reap value from EP methods. Thus, we have a common anchor with which we can approach a given EP method during assessment. Some part of the PCM system and monitored system must ultimately be the cause for a design choice or assumption taken in an EP method. Hence, the system perspective allows us to integrate the apparently disparate research in EP methods.}
To integrate and assess existing work on EP methods with the aim of facilitating knowledge sharing and deriving open challenges and research directions, this survey is guided by the following research questions: 

%\noindent\textbf{RQ1} How can existing event prediction methods be classified? \\
\noindent\textbf{RQ1} How can existing event prediction methods be assessed in a general, consistent, and \janikrev{system-oriented} manner? \\
\textbf{RQ2} To what extent do existing event prediction methods fulfill assessment criteria? \\
\textbf{RQ3} Which open challenges and research directions remain in the field of event prediction? 

\subsection{Research methodology}
\label{ssec:methodology}

\janikrev{The goal of answering \textbf{RQ1--RQ3} is to provide a comprehensive, consistent, and integrative overview of state-of-the-art EP across application domains and research areas and to facilitate knowledge sharing. 
%The two main challenges of achieving our underlying goal are as follows. On the one hand, 
Assume that a set of prediction goals $g_1, g_2, \ldots$ results from heterogeneous monitored systems and desired future states (cf. \autoref{fig:overview}). A set of different EP methods $m_1, m_2, \ldots$ is proposed for an EP goal. If EP method $m_i$ can be applied to continuously anticipate prediction goal $g_j$, we say that method $m_i$ is \emph{applicable to} $g_j$. An ideal situation for knowledge sharing would be to have a common terminology for equivalent ``applicable to'' relations between methods and goals. The literature analysis in \cite{zhao_event_2021} underpins that currently we do not have an ideal situation for knowledge sharing. Consequently, the first challenge for the survey at hand is to collect and resolve different \janikrr{conceptual models, implicit assumptions, and terminology} for equivalent prediction goals and corresponding methods. The second challenge is to integrate, assess, and present the vast amount of existing work such that our research questions are answered concisely.}

\begin{figure}[htb!]
    \centering
    \includegraphics[width=0.9\linewidth]{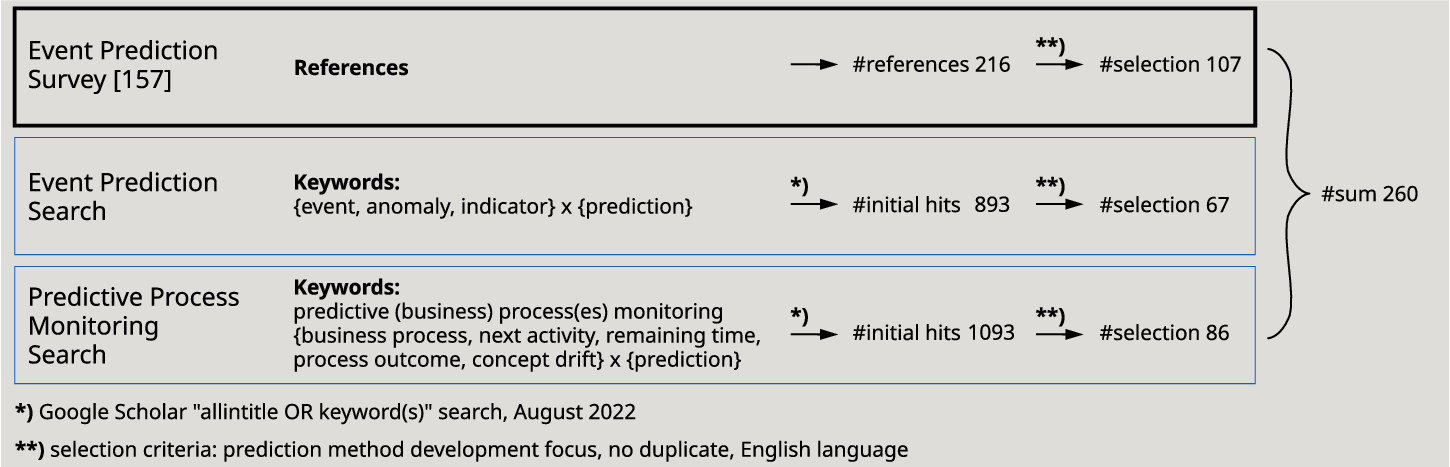}
    \caption{Methodology for compiling literature.}
    \label{fig:methodology}
    \Description{Illustration of the applied research methodology for compiling the literature of this survey. The existing survey \cite{zhao_event_2021} is used as a baseline with 216 references accompanied by two areas for literature search with a list of respective keywords: event prediction and predictive process monitoring. In the case of event prediction 893 articles and in the case of predictive process monitoring 1093 articles are found. We applied the same filter to each of the three initial literature compilations that requires a focus on prediction method development and English language and excludes duplicates. The result is the selection of 260 articles covered in this survey.}
\end{figure}

\janikrev{To overcome the challenge of different \janikrr{conceptual models, implicit assumptions, and terminology},} 
%we follow a twofold methodology for selecting relevant articles in the EP literature. W
we take the survey in \cite{zhao_event_2021} as a baseline on existing work. \janikrr{We} extend this baseline by searches in two directions: \janikrev{EP and PPM. The first direction aims to uncover EP method research that is missing in the baseline. As we have observed that EP research in the application domains engineering systems and cyber systems tend to use ``indicator'' or ``anomaly'' instead of ``event'', we have added both as keywords. The second direction aims to uncover missing EP method research mainly from the business application domain. The research area of business process management uses different terminology for EP, e.g., PPM or next activity prediction. We include all keywords that we have observed during analysis of EP research in business process management. Also, we cross-check our keywords with the keywords used in a recent survey of EP in the business domain \cite{rinderle-ma_predictive_2022}.} The methodology is illustrated in \autoref{fig:methodology} %($\usym{2780}$ literature compilation) 
with keywords for searches and the results. 

Given the \janikrr{2202} search results, we apply the same filter denoted by **) in \autoref{fig:methodology} %($\usym{2780}$) 
on each stream of literature resulting in selections that represent significant advancements in prediction method development. \janikrr{Here, significant advancement means }work that focuses on novel EP methods instead of only applying a classical machine learning algorithm to a new \janik{application domain} or focusing on domain-specific feature engineering. Lastly, we take the union of the three selections yielding 260 articles covered in this survey. The selection of 260 articles is classified \janikrr{into an integrated taxonomy} and \janikrev{reported in Section B in the} \janikrev{supplementary material}\footnote{\label{supplementary}\janikrev{The supplementary material consists of the supplemental online-only appendix Section A - F and our online repository \url{https://gitlab.com/janikbenzin/event_prediction_survey} for machine-readable material.}}. These articles span a wide array of different application domains each having a large enough absolute article count for proper representation (cf. Table \ref{tab:domain}). 

%\janikrev{In order to overcome the challenge of how to approach the review and assessment of existing literature, 
\janikrev{We combine an \emph{integrative review} \cite{whittemore2014methods} with an \emph{umbrella review}  \cite{whittemore2014methods} and the systems perspective \cite{bertalanffy1968general}. The integrative approach provides an overview of state-of-the-art EP through a qualitative research approach. 
The final selection of $260$ articles contains $34$ surveys from disparate research areas (cf. \autoref{fig:surveys}) which are taken as an initial yardstick for review and assessment in order to be consistent with existing work (cf. Figure 1 in the supp. material\supplementary). Synthesizing existing surveys from disparate research areas constitutes an umbrella review. 
%Taken together, we mix an integrative review with an umbrella review. 
Taking a systems perspective supports our mixed review by providing a mental \cite{bower1990mental} and reference model \cite{thomas2005understanding}. 
%Systems theory \cite{bertalanffy1968general} applied to event prediction aims to understand event prediction methods by simultaneously studying their environment. The environment of a method includes the monitored system, observer, input component, output component, and the user, comprised in a PCM system (cf. \autoref{fig:overview}). Thus, event prediction methods are applied in decision support systems (cf. \autoref{sec:intro} and \autoref{sec:arch}). 
%Moreover, the environment defines the purpose, semantics, and requirements imposed on the method (cf. \autoref{sec:req}). For example, the monitored system is related to an application domain (e.g., traffic observation in \autoref{fig:overview}) and determines what an event is and what it means. 
To this end, the PCM system explains the differences between EP methods and supports understanding different methods through a mental model. Also, it acts as a reference for aggregating requirements (cf. \autoref{sec:arch}) and as a tool for applying EP methods in real-world (cf. \autoref{sec:realizations}). }
%In the following, we present our analysis and integration approach resulting from our review type and perspective.}
%Additionally, it guides the realization of event prediction methods in the real-world (cf. \autoref{sec:realizations}).
%Through the systems perspective inherent in the PCM system, differences in event prediction methods are related to the respective system component. 

\janikrev{%For the classification ($\mapsto$ RQ1), we identify existing taxonomies in related surveys. The identified taxonomies are merged (cf. step 2 in Figure 1 in the appendix) into a single taxonomy (cf. \autoref{ssec:classification}) that is maintained throughout the whole review and assessment. The main challenges of merging taxonomies are the size of the merged taxonomy, different scales to distinguish types and different terminology for equivalent concepts. We choose to present a part of the overall taxonomy in \autoref{ssec:classification} and refer to our \janikrev{online repository}$^\text{\ref{supplementary}}$ for the overall taxonomy. 
For the general, consistent, and systematic assessment ($\mapsto$ RQ1), we identify existing requirements in the related surveys and merge them into an initial set of requirements (cf. Figure 1 in the supp. material\supplementary)%$\text{requirements}_0$ in \autoref{fig:methodology} $\usym{2781}$)
. Main challenges for merging requirements are different functional granularities and terminology for similar functionalities.}
\janikrev{We also encounter the challenge that identified requirements do not necessarily relate to the EP method itself, e.g., data requirements as identified in \cite{rinderle-ma_predictive_2022}. The system perspective helps us overcome this additional challenge. 
%The PCM system can relate all requirements to the respective components of a decision support system. The event prediction method is placed as the prediction component into the PCM system (cf. \autoref{fig:overview}). 
Similar to the initial set of requirements, the initial PCM system (cf. Figure 1 in the appendix%$\text{PCM system}_0$ from step 3 in \autoref{fig:methodology} $\usym{2781}$
) may not consistently, generally, and systematically represent all articles. Therefore, we apply the following analysis cycle (cf. Figure 1 in the supp. material) %\autoref{fig:methodology} $\usym{2781}$ 
for each of the 260 selected articles to overcome the challenge of uncovered requirements, \janikrr{different conceptual models and implicit assumptions.}}

\janikrev{The analysis cycle for current article $i$ begins by its classification into the taxonomy. Then, the article is assessed with the current set of requirements. Potential issues during assessment, e.g., some characteristics of the article are not represented by the requirements yet or imply to split a requirement into two separate requirements, determine the refinement of requirements. As the PCM system shall represent all requirements and generalize over all EP methods and their applications, it is also refined for every article. \janikrr{By placing the EP method into the PCM system and iteratively comparing it to already analyzed methods, we are able to uncover implicit assumptions. Additionally, we are able to map components of different conceptual models into an equivalent system context.} Last, the assessment of the previous articles is updated to represent the refined set of requirements. All in all, the analysis cycle for all 260 articles yields the PCM system and final set of requirements as presented in \autoref{sec:arch}.}

\subsection{Related Surveys}
\label{ssec:related}

%Due to its integration aim, this survey is related to different existing surveys. 
\janikrev{The article selection contains $34$ surveys.
%from disparate areas which are synthesized through an umbrella review. 
Some of these surveys %In the following, we refer to further surveys that 
focus on technical details of prediction methods in general and on specific problem formulations. 
\cite{zhao_event_2021} is taken as a yardstick for EP methods and
%Since the scope in \cite{zhao_event_2021} is more narrow and focused on the EP method itself, \cite{zhao_event_2021} 
presents the different methods in more detail when compared to the survey at hand.}
%Thus, the interested reader is recommended to read \cite{zhao_event_2021} as supplementary material. 
\janikrev{\autoref{fig:surveys} compares the survey at hand with the 34 related surveys from a methodological perspective. This further differentiates the survey at hand from existing ones and aims at highlighting its uniqueness and comprehensiveness. \janikrr{16 related surveys do not share any distinctive feature with the survey at hand. For example, \cite{zhao_event_2021} focuses more on classifying existing methods and gives a technical overview on approaches. Similar to 18 related surveys, }this survey is \emph{assessing} as it assesses the relative qualities of the EP methods either qualitatively or quantitatively. As outlined in \autoref{ssec:methodology}, this survey also combines an \emph{integrative review} \cite{whittemore2014methods} with an \emph{umbrella review} \cite{whittemore2014methods} and a systems perspective \cite{bertalanffy1968general}. \janikrr{We subsume the systems perspective and the integrative approach under \emph{system-integrative} (cf. \autoref{fig:surveys}).} 
%We base the classification of surveys as depicted in \autoref{fig:surveys} on these categories. 
A single related surveys %are assessing and one of these surveys 
is additionally system-integrative.} \janikrr{None of the surveys is an umbrella review. Hence, the survey at hand is the only one that aims at unifying existing EP method research in disparate areas by taking both research papers and surveys into account.}
%and 16 surveys cannot be classified in any of the categories. }
%A survey is system-integrative, if it is integrative using a system. The system perspective integrates application domains and disparate research areas into the PCM system (cf. \autoref{fig:overview} and \autoref{ssec:methodology}). The PCM system along with its requirements is a schematic for realizing the value of event prediction methods across research areas and application domains. System-integrative is a unique characteristic of our survey. A survey is user-oriented, if it takes a perspective on the event prediction methods that can represent the user of the method in the real-world. \cite{rinderle-ma_predictive_2022} is only other survey that is user-oriented through a similar systems perspective as our survey. Next, we provide a more fine-grained discussion of related surveys structured by the three pillars.

\begin{table}[ht!]
\resizebox{0.8\linewidth}{!}{\begin{tabular}{llccc}
\toprule
Area & Related surveys & Assessing & System-integrative & Umbrella   \\
\midrule
EP & \cite{zhao_event_2021,bhattacharjee_survey_2021,park_review_2021,jamaludin_systematic_2020,gmati_taxonomy_2019,molaei_analytical_2015,mehrmolaei_brief_2015} & \xmark & \xmark & \xmark \\
PPM & \cite{kamala_process_2022,di_francescomarino_predictive_2022,stierle_bringing_2021,wolf_framework_2021,spree_predictive_2020,lamghari_predictive_2019,marquez-chamorro_predictive_2018,di_francescomarino_predictive_2018,leontjeva_complex_2015} &  \xmark &  \xmark &  \xmark \\
EP & \cite{heyard_validation_2020,tama_empirical_2020,hoiem_comparative_2020,rebane_investigation_2019,tama_empirical_2019} & \cmark &  \xmark &  \xmark \\
PPM & \cite{kratsch_machine_2021,kappel_evaluating_2021,neu_systematic_2021,rama-maneiro_deep_2021,tax_interdisciplinary_2020,ogunbiyi_comparative_2020,weytjens_process_2020,harane_comprehensive_2020,tama_empirical_2019,verenich_survey_2019,teinemaa_outcome-oriented_2019,metzger_comparing_2015} & \cmark  & \xmark  &  \xmark\\
EP  & Not available & \cmark &  \cmark&  \xmark \\
PPM  & \cite{rinderle-ma_predictive_2022} & \cmark & \cmark &  \xmark\\
Any & Survey at hand & \cmark & \cmark & \cmark \\
\bottomrule
\end{tabular}}
\caption{\janikrr{Comparison of our survey with related surveys through three distinctive survey characteristics.}}
    \label{fig:surveys}
    \Description{Illustration of how related surveys are placed in terms of our survey's distinctive characteristic dimensions.}
\end{table}

\janikrev{For synthesizing the $34$ identified surveys from disparate areas, 
%we use the pillars depicted in \autoref{fig:methodology}.} Starting 
we start with the literature covered by EP survey \cite{zhao_event_2021} (cf. \autoref{fig:methodology}).} %surveys the literature on event prediction with the aim of a %systematic taxonomy and summary of existing approaches and application domains as well as standardizing event prediction evaluation for facilitating future development of event prediction. Considering the scope of \cite{zhao_event_2021}, our survey focuses 
\janikrev{By contrast to \cite{zhao_event_2021}, we focus on EP instead of only the EP method} (resulting in 107 of the 216 references) and extend the scope in the direction of uncovered EP method research (67 added articles) and uncovered PPM research (86 added articles). Aside from the scope, \cite{zhao_event_2021} focuses on the techniques to solve the identified EP research problems, e.g., regression. 
%point process and survival analysis technique for solving the problem of predicting a continuous point in time of future events. 
Hence, the existing approaches are classified and summarized along inherent properties of the proposed solutions. 
%With respect to classifying existing approaches, 
This survey extends the classification in \cite{zhao_event_2021} by abstracting from the respective technique and integrating taxonomies proposed in other surveys \cite{gmati_taxonomy_2019,di_francescomarino_predictive_2018,marquez-chamorro_predictive_2018,rinderle-ma_predictive_2022} \janikrr{(cf. Section B in supp. material\supplementary)}. With respect to summarizing existing approaches, our survey extends the summary in \cite{zhao_event_2021} by summarizing existing work along the requirements for EP methods that we elicit through a systems perspective, i.e., our summary takes an external, systems perspective on EP compared to the internal, methodological \janik{perspective} of \cite{zhao_event_2021}. Furthermore, the requirements are used to assess existing EP methods, resulting in a comprehensive assessment that further extends \cite{zhao_event_2021}. 

Overall, we consider eleven EP surveys \cite{bhattacharjee_survey_2021,park_review_2021,heyard_validation_2020,tama_empirical_2020,hoiem_comparative_2020,jamaludin_systematic_2020,rebane_investigation_2019,tama_empirical_2019,gmati_taxonomy_2019,molaei_analytical_2015,mehrmolaei_brief_2015} (cf. \autoref{fig:surveys}). 
Each of them has a certain scope, e.g., \cite{mehrmolaei_brief_2015,molaei_analytical_2015} focus on time-series data, \cite{tama_empirical_2019,tama_empirical_2020} focus on EP of business processes, and \cite{bhattacharjee_survey_2021} on unstructured text events. When compared to the survey at hand, on top of a narrow scope, the eleven surveys neither take a system perspective to generally place EP method in a context, nor elicit requirements for assessing the methods, nor comprehensively assess existing work. 
\cite{gmati_taxonomy_2019} proposes a taxonomy for EP methods with a concept of domain-specificity that is integrated in our taxonomy. 

We consider $22$ PPM surveys \cite{rinderle-ma_predictive_2022,kamala_process_2022,di_francescomarino_predictive_2022,kratsch_machine_2021,stierle_bringing_2021,kappel_evaluating_2021,neu_systematic_2021,wolf_framework_2021,rama-maneiro_deep_2021,tax_interdisciplinary_2020,ogunbiyi_comparative_2020,weytjens_process_2020,spree_predictive_2020,harane_comprehensive_2020,lamghari_predictive_2019,tama_empirical_2019,verenich_survey_2019,teinemaa_outcome-oriented_2019,marquez-chamorro_predictive_2018,di_francescomarino_predictive_2018,metzger_comparing_2015,leontjeva_complex_2015}  (cf. \autoref{fig:surveys}).
Similar to the reviewed EP surveys, the scope of the PPM surveys is limited. While the scope of EP surveys is heterogeneous, PPM surveys exclusively focus on next events (e.g., \cite{tama_empirical_2019}) or key performance indicators of business processes (e.g., \cite{verenich_survey_2019}). Only \cite{rinderle-ma_predictive_2022} takes a systems perspective for eliciting requirements and assessing existing PPM methods. 
%The scope in \cite{rinderle-ma_predictive_2022} includes compliance monitoring (CM) (cf. Table \ref{tab:abb}), i.e., the ability to continuously monitor the compliance status of a business process with respect to a set of compliance constraints, to investigate the potential benefits of combining CM and PPM with respect to the established CM functionalities framework \cite{DBLP:journals/is/LyMMRA15}. The investigation result is PCM as a combination of PPM and CM from a systems perspective. Existing work on CM and PPM is assessed with respect to an extended CM functionalities framework with a particular focus on prediction. 
Our survey generalizes the PCM concept to EP, develops a conceptual PCM system, and extends as well as integrates the requirements in \cite{rinderle-ma_predictive_2022,DBLP:journals/is/LyMMRA15} with the challenges in \cite{zhao_event_2021} and the requirements contained in the 260 articles of the existing work selection. 
\cite{di_francescomarino_predictive_2018,marquez-chamorro_predictive_2018} propose a similar concept of domain-specificity for classifying PPM methods that is generalized and integrated in our taxonomy. The distinction in \cite{rinderle-ma_predictive_2022} into methods that directly predict the goal of PPM and methods that indirectly predict the goal of PPM via an intermediate representation is generalized and integrated in our taxonomy. 
%For instance, the PPM goal is the late shipment status of online shop delivery packages. 
Predicting the late shipment status of an online shop delivery package, for example, is directly predicting the PPM goal and conceptually similar to the traffic status in the intersection example (cf. \autoref{fig:overview}). Predicting all future events pertaining to the delivery package and evaluating the shipment status on them is indirectly predicting the PPM goal and conceptually similar to predicting the number of vehicles on road segment $s_1$ to subsequently derive the traffic status in the intersection example.

If the time of the predicted event is of interest, %(\emph{occurrence time} prediction as introduced in \autoref{sec:req}), 
EP becomes a classical, supervised classification problem for which methods such as 
%Methods for classical classification problems, e.g.,
naive Bayes or logistic regression can be applied. 
%, are introduced in machine learning books. 
If the predicted events are represented as a time-series or necessitate time-series data (\emph{discrete} or \emph{continuous time} prediction), \cite{lim_time-series_2021,benidis_deep_2023} 
%provides an introductory and \cite{benidis_deep_2023} a technically advanced 
provide surveys based on deep learning, and \cite{masini_machine_2023} for other machine learning methods. If the predicted events are represented as spatio-temporal or necessitate spatio-temporal data (\emph{location} prediction), \cite{wang_deep_2022} provides a survey of deep learning methods and \cite{atluri_spatio-temporal_2019} for other machine learning methods. If the predicted event is an anomaly, EP methods are typically coupled with anomaly detection methods for which  \cite{boukerche_outlier_2021} provides a general overview and \cite{blazquez-garcia_review_2021} one for time-series data. \cite{neu_systematic_2021,rama-maneiro_deep_2021} provide surveys for categorical representation of the predicted event (\emph{semantic prediction}).

%All in all, our survey integrates various scopes of existing surveys and takes a system perspective for placing event prediction in a general context, eliciting requirements and assessing existing event prediction methods. 

\subsection{Contribution and Structure}
\label{ssec:contribution}

Taking a systems perspective on EP methods to integrate disparate areas of research results in the following four contributions of this survey:

\begin{itemize}
    %\item \textbf{Comprehensive classification of EP methods from a diverse range of application domains and disparate research areas.} EP method research spans areas like social sciences, healthcare, engineering, and computer science. Researchers at the intersection of these areas and data mining / machine learning propose a large collection of existing EP methods. This collection is classified using a taxonomy to support practitioners in finding suitable methods and researchers in developing new methods. 
    \item \textbf{Integration of disparate research areas through the generalized PCM system.} By taking a systems perspective, we identify an EP method as a component of the generalized PCM system. The generalized PCM system further integrates contributions from the various research areas on EP methods. The systems perspective emphasizes integration challenges, e.g., regarding interoperability of system components. \janikrr{We scrutinise the integration challenges through two realizations of the PCM system in manufacturing and healthcare (the healthcare realization is reported in Section E of the appendix).}
    \item \janikrr{\textbf{Structured requirement elicitation that encompasses the EP method within the PCM system.} The aim of integration through the PCM system is to elicit similar, yet broader requirements for EP methods. Integration through a set of general requirements facilitates the mutual exchange of method properties in general and of method properties that meet the requirements in particular. Without the broader view of the EP method in the PCM system, some, if not most, requirements are likely to be neglected or abstracted.}
    \item \textbf{Qualitative assessment of existing methods with respect to the requirements and thorough analysis of current research status.} A qualitative assessment of all existing methods gives an in-depth view on the current research status. The current research status is illustrated with examples from the literature, analysed with respect to potential dependencies between requirements and discussed in detail. 
    \item \textbf{Open challenges and discussion of future research directions.} The assessment result sheds light on blind spots that embody open challenges to be solved by future research. We identify five challenges and discuss the possible future research direction for each challenge. 
\end{itemize}

%The survey is structured into six sections: 
%of which the first four correspond to the four research questions respectively. To begin with, 
\autoref{sec:req} introduces a notion of an EP method and places
%gives the reasoning for placing an 
EP methods
%as an integral component 
into the generalized PCM system ($\mapsto$ RQ1). %\janikrr{Additionally, it showcases the PCM system}
%, presents the generalized PCM system in detail 
\autoref{sec:arch} explains the requirements for assessing EP methods ($\mapsto$ \janikrr{RQ1}). \autoref{sec:assessment} shows to what extent existing prediction methods fulfill requirements and pinpoints exemplary method properties that meet the respective requirement ($\mapsto$ \janikrr{RQ2}). \autoref{sec:challenges} highlights open challenges and distills research directions ($\mapsto$ \janikrr{RQ3}). %\janikrev{\autoref{sec:realizations} demonstrates how the PCM system can be applied in two application domains, i.e., healthcare and manufacturing.} 
Lastly, \autoref{sec:conclusion} concludes, \janikrev{presents our survey's impact} and identifies limitations of the survey. 
Note that we provide a supplemental online-only appendix and an online repository as supplementary material\supplementary. \janikrr{Amongst others, the supp. material features an integrated taxonomy of EP methods and a classification of all 260 EP methods into the integrated taxonomy.}

\section{A Notion and System Architecture for Event Prediction Methods}
\label{sec:req}
This section introduces a notion of the EP method (\autoref{ssec:problem}). 
\janikrr{It presents the rationale for the PCM system to embed EP methods into their system context (\autoref{ssec:arch}). It concludes with a demonstration of the PCM system in the manufacturing domain (\autoref{sec:realizations}).}
%It presents the rationale for our taxonomy of EP methods and the classification of EP methods based on our taxonomy (\autoref{ssec:classification}).

\subsection{Event Prediction Method Formulation}
\label{ssec:problem}

In the beginning, EP requires a goal or a set of goals that is either formulated by researchers during their motivation for advancement of EP methods or by practitioners during implementation and application of EP. Without the prediction goal, we do not know what we predict and lack a purpose for predicting. Prediction goals are as diverse as EP domains are. ``The traffic is green at $s_1$'' for the intersection example depicted in \autoref{fig:overview} is an EP goal. ``Reduce re-hospitalization and increase treatment outcome of heart failure patients'' \cite{duan_clinical_2020} is a set of EP goals. EP goals define the monitored system implicitly or explicitly and how we perceive various current and future states of the monitored system by differentiating desired or compliant from undesired or non-compliant states. In the intersection example, the goal implicitly defines the monitored system to be the depicted intersection and that green traffic is regarded as compliant, whereas black traffic is regarded as non-compliant. For hospital patients with previously diagnosed heart failures, the goal defines the monitored system to be hospital heart failure patients. Moreover, future states with reduced re-hospitalization numbers and improved treatment outcome statistics are desired in contrast to, e.g., states with constant re-hospitalization. 

\begin{definition}[Event Prediction (EP) Goal]
\label{def:goal}
An EP goal $g$ is a well-formed formula in some logic $G$ with atomic propositions over the domain of the monitored system that evaluates to either true or false given some predicted instantiation of its variables. If the EP goal $g$ evaluates to true, we say the monitored system future state is compliant, otherwise it is non-compliant. 
\end{definition}

For the sake of simplicity, we do not state the presented examples as formulas of a certain logic such as propositional logic or linear-time logic and point towards existing methods on translating natural language to formulas in, e.g., linear-time logic \cite{brunello_et_al:LIPIcs:2019:11375}. The two important points here are the existence of at least one EP goal and that it is stated in terms of atomic propositions over the domain of the monitored system (implicitly or explicitly). 

Based on the EP goal and the monitored system, we identify two kinds of data used for EP\footnote{We adopt the existing definition of the EP problem and method in \cite{zhao_event_2021} as close as our different perspective and scope allows.}, i.e.,  historical event data $Y_0 \subseteq \mathcal{T}^- \times \mathcal{L} \times \mathcal{S} $ and historical indicator data $X \subseteq \mathcal{T}^- \times \mathcal{L} \times \mathcal{F}_I$, where $ \mathcal{T}$ is the time domain, $ \mathcal{T}^- \equiv \{t | t \leq t_{\text{now}}, t \in  \mathcal{T} \} $ are all times up to the current time $t_{\text{now}}$, $ \mathcal{T}^+ \equiv \{t | t > t_{\text{now}}, t \in  \mathcal{T} \} $ are all times after the current time, $ \mathcal{L}$ is the location domain, $ \mathcal{S}$ is the event semantic domain and $ \mathcal{F}_I$ is the indicator feature domain that does not include the time and location domain \cite{zhao_event_2021}. Data types $Y_0$ and $X$ are generated by the observer (see \autoref{fig:overview}) through observing the real-world, e.g., the observer generates an event $y = (\text{``05.01.2023 13:00''}, \text{"$s_1$"}, \text{``Traffic = green''})$ by observing that two vehicles have stopped in front of the red light and an indicator $x = (\text{``05.01.2023 13:00''}, \text{"$s_1$"}, \text{"Workday = Yes"})$. This data is sent alongside the EP goal to the input component of the PCM system (cf. \autoref{fig:overview}) that maps the data to suitable features $ F \subseteq \mathcal{T}^- \times \mathcal{F}_Y  \times  \mathcal{F}_X$, where  $\mathcal{F}_Y$ is the domain of features based on historical event data and $\mathcal{F}_X$ is the domain of features based on historical indicator data, by means of preprocessing and encoding. The input component outputs the features and prediction goal to the prediction component that is realized by an EP method, e.g., $f = (\text{"05.01.2023 13:00"}, (1, 0), 1)$ for a one-hot encoded traffic value and identifying the workday by 1. 

\begin{definition}[Event Prediction (EP) Method \cite{zhao_event_2021}]
\label{def:prediction}
Given features $F \subseteq \mathcal{T}^- \times \mathcal{F}_Y \times \mathcal{F}_X$ and an EP goal $g \in G$ or set of goals, an EP method outputs predicted future events $\hat{Y} \subseteq \mathcal{T}^+ \times \mathcal{L} \times \mathcal{S}$ such that for each future predicted event $\hat{y} = (t, l, s) \in \hat{Y}$ for $t > t_{\text{now}}$ it holds that\janik{~\circled{C}} either the goal $g$ or the set of goals is itself a future predicted event or it can be evaluated on the future predicted events.
\end{definition}

In the traffic example, let us assume that we learn a neural network model that takes the most recent feature $f$ and outputs $\hat{y} = (\text{``05.01.2023 14:00''}, \text{``$s_1$''}, \text{``Traffic = black''})$ as the predicted future event (cf. \autoref{fig:overview}), i.e., the goal is itself the predicted event. Consequently, the future state of the traffic at road segment $s_1$ is predicted to be non-compliant such that an action by the user is required. 

Following the scope and aim as set out in \autoref{ssec:methodology}, we separate feature engineering from EP and focus on the EP method. We add condition ~\circled{C} on prediction goals. 
%that the event prediction method has an 
~\circled{C} expresses the benefit of predicting the future events and how we perceive the possible predicted future events, namely as compliant with our desired future state or as non-compliant. Defining EP methods with an explicit goal has three benefits. First, all existing work in our selection state a goal or set of goals in their motivation that expresses the benefits to users and how a specific predicted future event is perceived, e.g., \cite{duan_clinical_2020} predict future patient treatment outcomes in clinics for a proactive treatment effect analysis. Hence, including condition ~\circled{C} in the definition combines the \textsl{how} (i.e., the EP method) and the \textsl{why} (i.e., the prediction goal) for EP. Second, it links predicted future events to actions. Since an action aims to change a monitored system towards a desired state, linking actions to future predicted events requires a notion of a desired state. With prediction goals, the notion of desired states is defined by compliant or desired future states and by non-compliant or undesired future states. %Based on this notion, an action can aim for a desired future state of the monitored system. 
Third, condition ~\circled{C} emphasizes how we can decide whether a specific EP method is successful in predicting future events.

%Furthermore, condition ~\circled{C} captures the fact that not all prediction methods predict occurrence times, locations and semantics of future events simultaneously, as many focus on a single domain that is sufficient for the prediction goal.  \cite{diraco_machine_2023}, \janikrev{for example}, focuses on the time domain by predicting the occurrence of abnormal deformations for a single machine tool of a production facility, \cite{pirajan_towards_2019} focuses on the location domain by predicting the area that a particular crime will occur, and \cite{mehdiyev_multi-stage_2017} focuses on the semantic domain by predicting the future activity of a business process instance. 

If we want to fully understand EP in all of its heterogeneous aspects, multi-scale dimensions and overall complexity, we must achieve to grasp the set of systematic relationships of the system in which EP is conceptualized or implemented \cite{rice_leveraging_2021,elgin2007understanding,elgin2017true,kvanvig2003value,grimm2006understanding}. 
This feature of understanding emphasizes the need for a systems perspective on this topic. 
In general, considering the diverse nature of prediction goals throughout the various application domains, the type of system for EP must be a decision support system rather than an autonomous system, i.e., the user that decides what action is carried out is a human person (cf. \autoref{fig:overview} and \autoref{fig:framework}). For instance, deciding on the treatment plan of patients based on predicted future treatment outcomes must be done by physicians \cite{decamp_why_2019}.
Furthermore, the utility of EP critically depends on the lead time, i.e., the time from the prediction of an event to the true event time. 
This property necessitates a prediction at run-time and, thus, an online, stream-based conceptualization of the EP system \cite{salfner_survey_2010}.
The source streams have to be continuously monitored such that the lead time is as large as possible \cite{zhao_event_2021}. 
Altogether, we need a decision support system that continuously monitors and predicts with the aim of distinguishing non-compliant, action-demanding from compliant future states. 

To that end, we present a conceptual PCM system that entails these properties \janikrr{in the next section and operationalize the PCM system in \autoref{sec:realizations}. In \autoref{sec:arch}}, we elicit assessment requirements from the system architecture \janikrr{to asses existing EP methods in their ``PCM system readiness''}.

\subsection{Widening the Scope: Event Prediction Method in the PCM system}
\label{ssec:arch}

Before presenting the PCM system, we elaborate on how it applies to EP. This enables us to understand how multiple research areas work towards the same goal with more or less similar methods, yet different terminology. Moreover, we illustrate the community- and application-domain-spanning characteristic of EP. 

\begin{wrapfigure}[14]{R}{6cm}
  \centering
  %\vspace{-0.75em}
  \includegraphics[width=\linewidth]{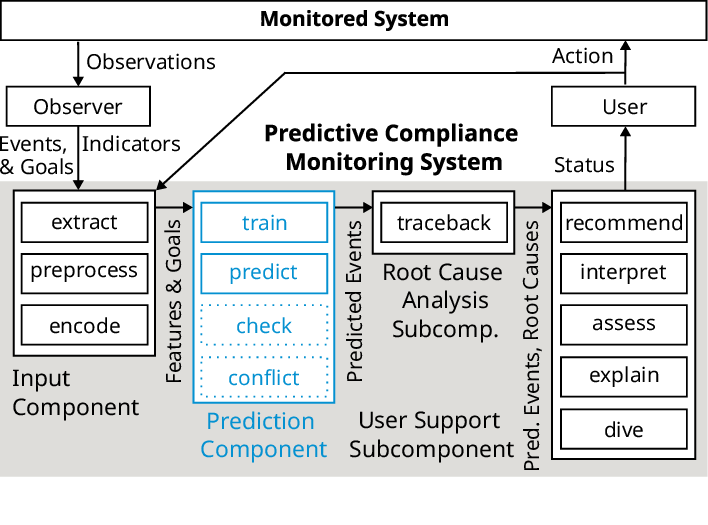}
  \caption{Generalized Predictive Compliance \\Mo\-ni\-toring System -- Conceptual and Structural View}
  \label{fig:framework}
  \Description{Illustration of predictive compliance monitoring system in more detail. The input component is refined to have extract, preprocess and encode functionalities, the prediction component is refined to have train, predict, potentially check and conflict functionalities, the output component is refined to have two subcomponents root cause analysis and user support. The root cause analysis component implements the traceback and the user support component implements recommend, interpret, assess, explain and dive functionalities.}
\end{wrapfigure}

In the field of business process management, a PCM system supports companies in monitoring and understanding the future compliance status of their business processes to act proactively towards a desired future compliance status \cite{DBLP:journals/is/LyMMRA15,rinderle-ma_predictive_2022}. To that end, a PCM system integrates research on compliance monitoring (CM) and PPM. CM aims to evaluate compliance constraints on the current and future events or key performance indicators of a business process \cite{DBLP:journals/is/LyMMRA15}, while PPM aims to predict the future events or key performance indicators of a business process \cite{di_francescomarino_predictive_2018,marquez-chamorro_predictive_2018}. By comparing CM, PPM, and the EP method definition (cf. \autoref{def:prediction}), we identify the following similarities: Compliance constraints are prediction goals, future key performance indicators of a business process correspond to future predicted events (cf. condition ~\circled{C} in \autoref{def:prediction}), PPM is an EP method, and the business process is a monitored system. If we generalize 
\begin{enumerate}
    \item the monitored business process and its events as a monitored system and its events, 
    %\item the predicted target of business processes as events,
    \item the prediction of the future compliance status as EP methods,
    \item the anticipated compliance status as the evaluation result of an EP goal,
    \item and the proactive actions as any action an user can take with the support of a decision support system,
\end{enumerate}   
then the generalized PCM system depicted in \autoref{fig:framework} is a conceptual model of decision support systems featuring EP. 
Not only does this generalization place EP inside a generic system necessary for understanding its merits more deeply, it also represents an integration of research on the various aspects of EP and monitoring into a common theme. Furthermore, the generalization of a PCM system from business process management to a decision support system for EP in general enables us to mutually transfer requirements and solutions between the respective research areas.
A PCM system contains the input, prediction, and output component (cf. \autoref{fig:overview}) which are refined in \autoref{fig:framework}. Moreover, a feedback cycle is added from the user to the input component that contains the action as a more direct and efficient way of accounting for the effects of derived actions onto the monitored system in the PCM system \cite{zhao_event_2021}.

\subsubsection{Input component.} 

The input component \emph{extracts} data from the observer, \emph{preprocesses} the data and \emph{encodes} it into the respective feature space required by the EP method (cf. \autoref{def:prediction}). In practice, the input component has to implement the various functionalities of \emph{data collection} proposed by the data management community, in particular data acquisition and data labeling \cite{roh_survey_2021}. Within data acquisition, the tasks of augmentation through data integration / data fusion are particularly important. \janikrr{For the case of input event data lacking an event notion}, the task of data programming or event extraction \cite{diba2020extraction} become additionally relevant. Considering that the observer is only an abstraction for multiple, heterogeneous data sources (cf. \autoref{sec:intro}), the input component's \emph{extract} functionality has to deal with challenges of raw data ingestion from various data sources \cite{meehan_data_2017}.

The input component's \emph{preprocess} functionality covers the transformation of ingested data to a common representation based on a series of rules \cite{jin_foofah_2017}, data cleaning \cite{ilyas_trends_2015}, schema matching for relational data \cite{rahm_survey_2001}, deduplication and record linkage \cite{elmagarmid_duplicate_2007,getoor_entity_2012,naumann_introduction_2010}, and, finally, canonical column selection for relational data \cite{stonebraker_data_2018}. In the case of non-relational data such as continuous, unlabelled sensor readings, schema matching is replaced by data programming or event extraction \cite{diba2020extraction} to define an event notion. In case of low-level events coming from the event source, the task of event abstraction \cite{van_zelst_event_2021} may be necessary. Since events from multiple sources may actually be the same event, these events have to be matched to finalize the data integration. 
As a last step, the \emph{encode} functionality maps the integrated data to the event $ \mathcal{F}_Y$ and indicator feature space $\mathcal{F}_X$. Here, it is important to follow the principles of feature engineering \cite{zheng2018feature}, while the resulting features are typically domain- and/or goal-specific. 

\subsubsection{Prediction component.} The prediction component is realized by an EP method that \emph{trains} a prediction model and \emph{predicts} the future events. In \autoref{def:prediction}, the future event can either be \janik{(i)} the prediction goal or \janik{(ii)} an intermediate event on which the prediction goal can be evaluated. \janik{Case (i)} does not require the additional functionalities \emph{check} and \emph{conflict}. \janik{For case (ii)}, the evaluation of the prediction goal is realized by \emph{checking} the compliance of the intermediate event with respect to the prediction goal. If a monitored system is subject to a set of prediction goals, these goals may be in conflict \cite{DBLP:journals/is/LyMMRA15}. The PCM takes this possibility into account when checking the set of prediction goals on the intermediate event and includes this information in its output. 

\subsubsection{Output component.} The output component is refined into two subcomponents: the \textbf{root cause analysis component} and the \textbf{user support component} (cf. \autoref{fig:framework}). \janikrr{The refinement} underlines the need for model transparency and interpretability (in short explainability), accountability, and prescriptiveness \cite{DBLP:journals/is/LyMMRA15,rinderle-ma_predictive_2022,zhao_event_2021}. The ability of the PCM to \emph{trace} the non-compliance of the future events \emph{back} to the root cause(s) contributes to an explainable and actionable prediction. The prediction with root cause analysis is explainable, because a causal relationship between a root cause and its anticipated result exists \cite{rice_leveraging_2021}. Furthermore, root cause analysis makes the prediction actionable, since the user can derive countermeasures for non-compliant future events through acting on the root causes. In engineering and cyber systems, i.e., IT systems, root cause analysis is typically developed for explaining the root causes of machine faults or system anomalies to the operation manager \cite{e_oliveira_automatic_2022,ma_big_2021}. 

Finally, the PCM system supports the user in understanding the results and acting on them. To that end, the user support subcomponent \emph{recommends} actions and countermeasures for non-compliant predicted future events \cite{alagoz_markov_2010,park_action-oriented_2022} and \emph{interprets} the impact of the proposed actions on the predicted future events \cite{grisold_adoption_2020}. \janikrr{Further, it} \emph{assesses} the overall uncertainty that remains throughout all prior functionalities \cite{hullermeier_aleatoric_2021}. Separately showing all the results to users may quickly overwhelm, distract from major points, and hinder understanding. \janikrev{Hence,} the gist of the results are \emph{explained} to them. In \autoref{sec:challenges}, we discuss a potential approach to this challenging functionality. If more detailed information is necessary for understanding or analysing, the user can \emph{dive} into the respective results. 

%All in all, principles of systems engineering applied to the design of a conceptual decision support system featuring EP result in a new perspective on EP \cite{blanchard1990systems}. 
This new systems perspective encompasses a holistic view on the many facets of EP, yet is simple enough to guide us in our understanding of functionalities surrounding the actual prediction that are necessary to hone its value. \janikrev{Eventually, the PCM system depicted in \autoref{fig:framework} can be seen as an approach and system blueprint for the realization of real-world applications, comparable to the \emph{business model canvas} for business model creation \cite{osterwalder2010business}. Two PCM system realizations from manufacturing and healthcare are presented in the next section and in Section E of the supp. material\supplementary.}

%\stefanierev{\subsection{Applying the PCM System in Manufacturing and Health Care}}

\janikrr{
\subsection{Realization of a PCM System in Manufacturing}
\label{sec:realizations}}

\begin{figure}[htb!]
  \centering
  \includegraphics[width=0.8\linewidth]{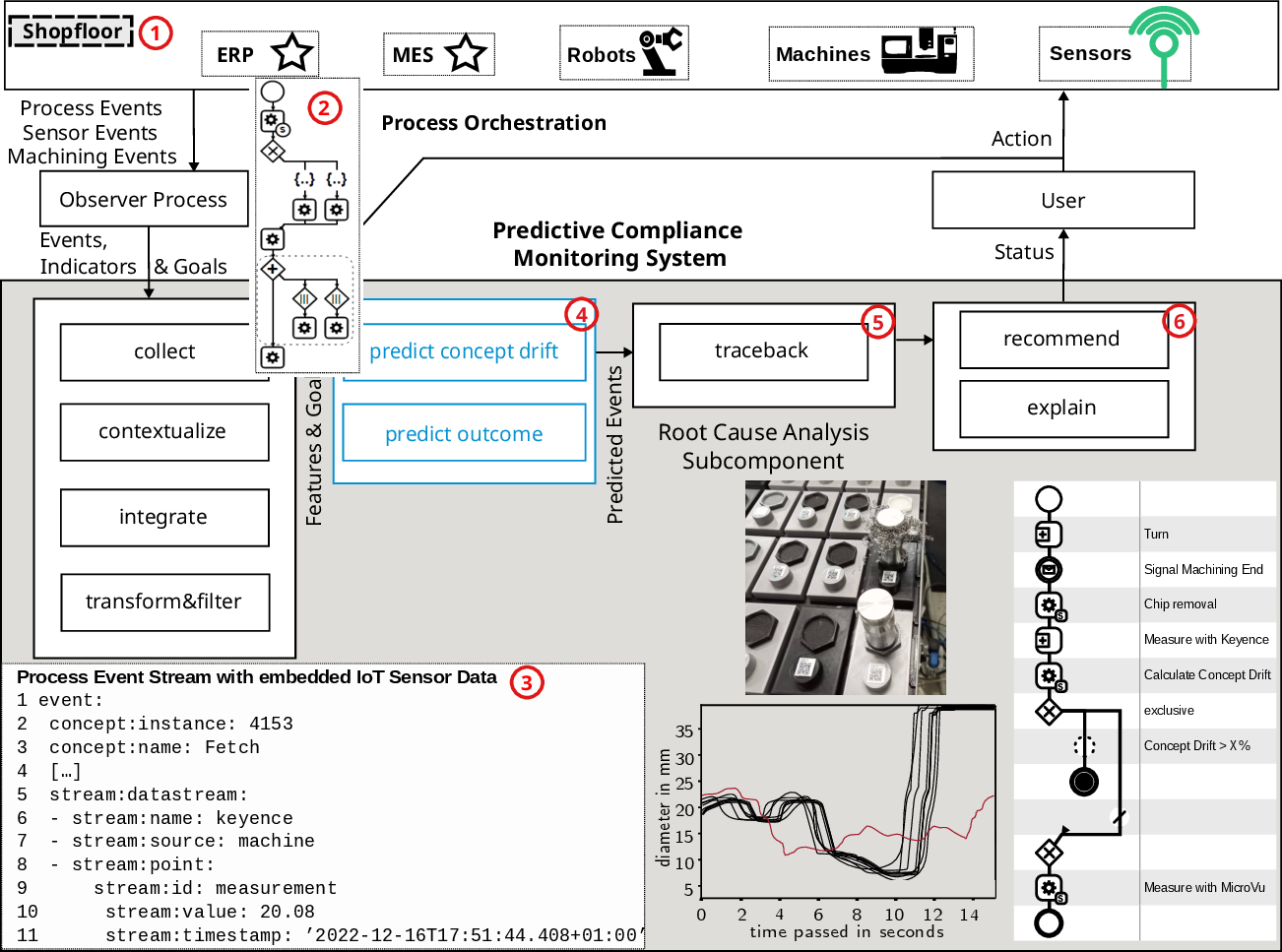}
  \caption{PCM Approach in Manufacturing: Predicting and Explaining Concept Drifts}
  \label{fig:framework_manufacturing}
  \Description{Illustration of the PCM Approach in Manufacturing through operationalizing the conceptual and structural view on the PCM system (cf. \autoref{fig:framework}) with the system component's instantiations in practice. The monitored system is instantiated by the shopfloor with event sources like robots and machines. The observer becomes an observer process in the form of a process orchestration. The input component is instantiated to collect, contextualize, integrate, and transform\&filter the raw data. The observer process instantiation allows to embed IoT sensor data directly into the event. Next, the prediction component consists of predict concept drift and predict output followed by the root cause analysis component with a traceback through concept drift analysis. Lastly, the output component recommends and explains the resulting predictions and concept drifts to the user.}
\end{figure}

\janikrr{In this section, we show how the PCM approach (cf. \autoref{fig:framework}) can be used in order to realize EP projects at the example of manufacturing. \textsl{``Manufacturers need to know what happens next and what actions to take in order to get optimal results. It is a challenge to develop advanced analytics techniques including machine learning and predictive algorithms to transform data into relevant information for gaining useful insights to take appropriate action''} \cite{DBLP:conf/wecwis/ThalmannMSHSPW018}. This means that several event sources in a cyber-physical environment are to be leveraged to predict drifts and outcomes in a production process. %, i.e., the PCM approach acts as a system blueprint similar to the \emph{business model canvas} for business model creation \cite{osterwalder2010business}. 
%By widening the scope from a single EP method to the PCM system, particular concerns and requirements of domain experts can be naturally placed and represented accordingly. %The healthcare example integrates EP methods from literature and shows how researchers can leverage the PCM approach to integrate and develop their methods in a specific application context. 
%The manufacturing example illustrates how PCM  can be applied to integrate event sources in a cyber-physical environment and to leverage prediction methods to predict drifts and outcomes in a production process.
}
\janikrr{Figure \ref{fig:framework_manufacturing} illustrates how the PCM system can be applied to help achieving these goals in manufacturing. The described use case brings together requirements and settings from real-world projects as described in \cite{DBLP:books/sp/21/PaukerMRE21,DBLP:conf/bpm/StertzRM20,DBLP:journals/fi/ManglerGMEBBRAB23,DBLP:conf/icsoc/EhrendorferMR21,DBLP:conf/caise/ScheibelR22,DBLP:conf/caise/ScheibelR23}.}

\janikrr{The situation faced concerns the introduction of a cyber-physical system \textsl{``tracking new orders from an Enterprise Resource Planning (ERP) system, sourcing materials and scheduling production resources from a Manufacturing Execution System (MES), and collecting data from all machines, robots, and other equipment used on the shop ﬂoor''} \cite{DBLP:books/sp/21/PaukerMRE21} (cf. Fig. \ref{fig:framework_manufacturing} \ding{172}). The integration is realized through process orchestration \cite{DBLP:books/sp/Rinderle-MaMR24} acting as \textsl{observer process} (cf. Fig. \ref{fig:framework_manufacturing} \ding{173}) that connects to the different systems, machines, robots, and sensors using interfaces such as OPC UA and collects data through the Siemens S7 communication protocol \cite{DBLP:books/sp/21/PaukerMRE21}. The log snippet depicted in Fig. \ref{fig:framework_manufacturing} \ding{174} shows process event \texttt{Fetch} and the associated sensor stream for keyence measurements in the DataStream XES format \cite{DBLP:journals/fi/ManglerGMEBBRAB23}.  }

\janikrr{Prediction goals in manufacturing (cf. Fig. \ref{fig:framework_manufacturing} \ding{175}) comprise prediction of drifts in the prescribed process behavior \cite{DBLP:conf/bpm/StertzRM20} and in the process data \cite{DBLP:conf/caise/StertzR19}. In addition, the process outcome \cite{DBLP:conf/icsoc/EhrendorferMR21} such as the quality of the produced part and monitoring the decision rules that drive the routing of the process are typical prediction goals \cite{DBLP:conf/caise/ScheibelR22,DBLP:conf/caise/ScheibelR23}. For example, monitoring a temperature measurement exceeding a certain threshold for three times in a row can be crucial for quality assurance. Taking the prediction of concept drift \cite{DBLP:conf/bpm/StertzRM20} as another example, the approach combines predicting the next event in a process stream with the prediction of sensor data values, e.g., the data stream representing the diameter measurement of the produced parts. EP methods can predict drifts in the diameter measurements. These drifts can he analyzed and used as basis to recommend (cf. Fig. \ref{fig:framework_manufacturing} \ding{177}) an adaptation in the production process. The occurrence of chips on the part results in diameter drifts and, hence, a new task \texttt{chip removal} is executed. The approach can be also used for explanation (cf. Fig. \ref{fig:framework_manufacturing} \ding{177}), i.e., if a concept drift occurs, the root cause can be traced back to the diameter measurements (cf. Fig. \ref{fig:framework_manufacturing} \ding{176}). }

\janikrr{Utilizing the PCM approach for describing the manufacturing scenario yields an overview of the entire pipeline: from the data sources to the prediction methods, insights, and actions to be taken. It also shows which challenges are still open in realizing the pipeline in its entirety. First of all, a systematic way of combining EP methods and PPM methods is missing. Moreover, the presentation and explanation of insights to the user is still in its infancy, requiring more research to keep users in and on the loop. } 
%\todo{@Stefanie: Maybe add more challenges for integration specific to the manufacturing use case + reference interop book?}

\noindent\textbf{Summary}: Using the PCM approach in the manufacturing case underpins the gap between requirements that are met by existing approaches and requirements that are not met, but highly relevant in EP projects, i.e., data quality, lifecycle support, and keeping humans in the loop by, i.a. advanced root cause analysis.
\hfill \break

\section{Comprehension of Event Prediction Methods through the Generalized PCM System}
\label{sec:arch}

\begin{figure}[htb!]
  \centering
  \includegraphics[width=0.75\linewidth]{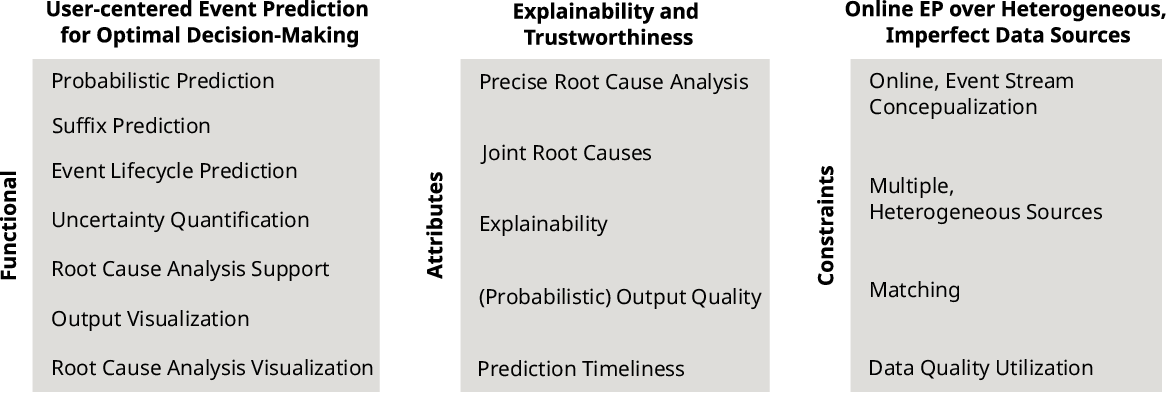}
  \caption{Requirements engineering result for EP methods in the context of PCM systems categorized by functional requirements, attributes and constraints \cite{glinz_non-functional_2007} and annotated by \janikrr{the underlying theme.}}
  \label{fig:req}
  \Description{Illustration of the 16 requirements engineered based on the architecture in \autoref{fig:framework} and the functionalities in \cite{rinderle-ma_predictive_2022}: Probabilistic Prediction, Suffix Prediction, Event Lifecycle Prediction, Uncertainty Quantification, Root Cause Analysis Support, Output Visualization, Root Cause Analysis Visualization, Precise Root Cause Analysis, Joint Root Causes, Explainability, Probabilistic Output Quality, Prediction Timeliness, Online Event Stream Conceptualization, Matching, Data Quality Utilization and Multiple, Heterogeneous Sources.}
\end{figure}

Based on the refined understanding of EP 
%through its conceptualization 
as a component of PCM systems, we take the systems perspective one step further \janik{by eliciting requirements for EP methods from existing surveys \cite{zhao_event_2021,rinderle-ma_predictive_2022} and iteratively refining them during the assessment of the 260 articles in our selection}. Aside from advancing our account of EP, the requirements serve as an assessment scheme for assessing the selection of existing work. \janik{Following a concern-based taxonomy of requirements \cite{glinz_non-functional_2007}, the 16 EP method requirements in \autoref{fig:req} are classified into seven functional and nine non-functional requirements in the form of five attributes and four constraints}. Attributes specify qualities of the method, whereas constraints limit the solution space from which methods can be instantiated. \janikrr{To emphasize how meeting the requirements impacts the PCM system, we identify an underlying theme for each group of requirements. The seven functional requirements  enable optimal decision-making \cite{alagoz_markov_2010} and center around the user through visualization \cite{mohammed_big_2022}. The five attributes represent the need for an explainable and trustworthy PCM system \cite{barredo_arrieta_explainable_2020,burkart_survey_2021}. Lastly, the four constraints specify a PCM system with online EP that continuously predicts future events from heterogeneous, imperfect data sources \cite{rinderle-ma_predictive_2022}.}
%Furthermore, each requirement is assigned to one of four specification targets\todo{possibly remove these}. 

%\janikrr{First,} the prediction method's output refers to the properties of predicted future events (I). \janikrr{Second,} understanding the prediction method refers to properties or qualities of the method that help to grasp the systematic relationships between the prediction and its input through, for example, visualization \cite{mohammed_big_2022,ali_big_2016} (II). \janikrr{Third,} evaluating the prediction method refers to the qualities that presented for evaluation (III). \janikrr{Fourth}, the prediction method's input refers to what properties of the input were considered in developing the method (IV). 

\janikrr{\textbf{Optimal decision making and user-centeredness.}} Considering that the real-world is complex, chaotic, and imperfect, a deterministic prediction of future events idealizes and keeps important nuances in the future events from the user \cite{zhao_event_2021,rice_leveraging_2021,rinderle-ma_predictive_2022}. Therefore, a \emph{probabilistic prediction} that includes information on the likelihood of the predicted future events actually occurring is required. If we want to enable the user to optimize decisions with respect to a comprehensive anticipation of the future, predicting the next event is not enough. A \emph{suffix prediction} specifies the need for anticipating more events following the next event, e.g., knowing of the next three heavy rainfalls instead of the next one, officials can more efficiently allocate resources. Instead of increasing the anticipated time horizon as suffix prediction does, \emph{event lifecycle prediction} refines the granularity of the anticipated event time horizon. Although events are typically conceptualized as real-world occurrences happening at a certain point of time, each real-world event has a duration and, thus, at least a \emph{start} and an \emph{end} point of time \cite{leemans_partial-order-based_2022,DBLP:journals/is/LyMMRA15}. If we do not predict the lifecycle of events, we are, for example, not aware of the time period affected by the event. \janikrr{For example, }our travel plans remain unchanged, if the traffic jam is likely to vanish by the time we arrive at its location. 

\emph{Uncertainty quantification} aims at reflecting the epistemic and aleatoric uncertainty of the method's output \cite{hullermeier_aleatoric_2021}. Epistemic uncertainty consists of model and approximation uncertainty and is the result of our lack of knowledge on the true relationships of the monitored system necessary for the design of a perfect prediction method. As such, it is reducible by further research. Aleatoric uncertainty is an irreducible uncertainty that captures the non-deterministic nature of real-world occurrences. Recently, method development acknowledged the existence of both types of uncertainty and incorporated their quantification in its output \cite{weytjens_learning_2022}. Uncertainty quantification can build trust for the prediction method by measuring confidence metrics \cite{zhao_event_2021} that accompany probabilistic outputs and improves the subsequent actions, as the user can decide to wait for more certain prediction results that are more likely to occur. 

Despite the fact that root cause analysis and visualization are separated from prediction in the PCM system (cf. \autoref{fig:framework}), it is important that the method itself enables these functionalities or already presents a proof of concept on how these functionalities can be done. Otherwise, it will be more difficult for researchers and practitioners to develop these functionalities given only the method itself. Thus, \emph{root cause analysis support} (RCA support), \emph{output visualization} and \emph{root cause analysis visualization} (RCA visualization) indicate whether the proposed method clearly defines how to achieve root cause analysis or visualization on top of the method respectively or already presents a proof of concept. \janikrr{In particular, the visualization functionalities focus on the user in the PCM system \cite{mohammed_big_2022}.}

\janikrr{\textbf{Explainability and trustworthiness.}} The first set of attributes are the specific qualities \emph{precise root cause analysis} (precise RCA), \emph{joint root causes} (joint RCs), and \emph{explainability} that are essential for an EP method, since this first set of attributes helps the user to quickly grasp the situation and act accordingly. The method should enable the precise identification of root causes and detect that multiple true root causes for different predicted future events may, in fact, be the same true root cause. As it is out of the scope of this survey to assess the full variety of how explainability can be achieved or conceptualized \cite{barredo_arrieta_explainable_2020,burkart_survey_2021} and there is no \janik{widely accepted definition \cite{rice_leveraging_2021,burkart_survey_2021,gao_going_2022}}, we limit the attribute to \janikrr{coarse-grained} distinction. \emph{Black box} models are intransparent, \emph{white box} models are transparent, and a combination of both exhibit both transparent and intransparent components \cite{burkart_survey_2021}. Taken together, these three qualities determine the accountability of the method to a large extent. If the user can explain the predicted future events as prerequisites to an optimal strategy for acting on the true root causes, the method and the user both should not be held accountable in case something goes wrong. 

The second set of attributes are performance attributes that evaluate the EP method itself. \cite{zhao_event_2021} presents how the output quality of an EP method is evaluated. Considering the requirement on a \emph{probabilistic output quality} (probabilistic quality), we further require a probabilistic evaluation, e.g., with the Brier score \cite{benedetti_scoring_2010} or some other appropriate scoring rule \cite{doi:10.1287/deca.2016.0337}. \emph{Prediction timeliness} pertains to the speed a method is able to predict. \janikrev{Prediction timeliness is particularly important in resource-constrained environments such as vehicle onboard systems \cite{akl_trip-based_2021} or multi-tenant cloud architectures with short customer response times \cite{singh_multi-objective_2022}.} It determines the lead time, \janik{but as a result of concept drift, the training time also has an effect on the lead time}. Due to changing environments such as the COVID-19 pandemic concept drift occurs and leads to outdated historic data \cite{rizzi_how_2022,zhao_event_2021,rinderle-ma_predictive_2022}. Consequently, the prediction model must be kept up-to-date. \janik{Having to update prediction models results in the training time of the method playing an additional role for a timely prediction, yet we do not know the significance of this role. Moreover, it is not clear what the optimal time for updating is.} Aside from these unknowns, training and prediction time comparisons are only reasonable in a fair benchmark that goes beyond the scope of this survey. Hence, prediction timeliness is not assessed in \autoref{sec:assessment}. 

\janikrr{\textbf{Online EP over heterogeneous, imperfect sources.}} \janikrr{To place EP methods into the PCM system, the following four constraints must be met.} First, \emph{online, event stream conceptualization} (online conceptualization) is a consequence of the need for monitoring and timely prediction \cite{rinderle-ma_predictive_2022,zhao_event_2021}. A method may fail to show how it can be applied to an event stream at run-time. There is a difference between online prediction based on an event stream and offline prediction based on an ex-post dataset, e.g., in terms of efficiently storing and updating already received events from the event stream \cite{cho_-line_2011}. \janikrr{Thus,} offline EP methods may be unsuitable for monitoring. 
Second, \emph{multiple, heterogeneous sources} (multiple sources) limit the method solution space to methods that are aware of or actively deal with \janik{the challenges of data integration} (cf. input component in \autoref{ssec:arch}). Although overcoming these challenges is the responsibility of the input component in a PCM system, the solution may affect the EP method, e.g., by necessitating multiple prediction models \cite{ramakrishnan_beating_2014} \janikrev{or multivariate problem formulations \cite{penalvo_sustainable_2022}}. 

Third, the challenge of \emph{matching} events from multiple data sources is emphasized as a separate constraint (cf. input component in \autoref{ssec:arch}). The matching strategy for the input should be consistent with the strategy for matching events in prediction goals to predicted future events and the strategy for matching events for evaluation purposes. Multiple matching strategies exist \cite{zhao_event_2021}, but may have to employ different equivalence notions that underlie the matching of events \cite{rinderle-ma_predictive_2022}. \janikrr{For example, }matching can be based on equivalent machine identifiers for multiple sensors attached to a machine in a production facility (cf. \autoref{sec:realizations}).
Fourth, \emph{data quality utilization} (data quality) limits the solution space to methods that not only acknowledge the existence of data quality issues such as missing data in their design, but also improve the prediction by actively exploiting the existence of data quality issues. For example, \cite{lu_self-supervised_2021} develops a self-supervised technique to fully utilize electronic health records that inherently lack a certain type of label. 

Conjointly, 15 of the 16 discussed requirements (excluding the \emph{prediction timeliness} requirement) serve as the assessment scheme for existing work on EP methods in \autoref{sec:assessment}.

\section{Assessment of Existing Event Prediction Methods}
\label{sec:assessment}
This section presents the analysis of $260$ papers from the literature compilation in \autoref{fig:methodology}. 
Each paper is analyzed with respect to the 15 requirements contained in the assessment scheme introduced in \autoref{sec:arch}, resulting in an overview for each requirement depicted in \autoref{fig:sum}.  
%The analysis of each paper is then aggregated along the requirements to give an 
%and aggregated along the concernefor each requirement as depicted in \autoref{fig:sum}. 
%Following a requirement-based view, this section begins with aggregate results in \autoref{ssec:aggregate} and continues with the three groups of requirements that are based on functional requirements in \autoref{ssec:functional}, attributes in \autoref{ssec:attributes} and constraints in \autoref{ssec:constraints}. 
The goal of the analysis is to understand the current status of EP methods from a systems perspective (cf. \autoref{sec:intro} and \autoref{sec:arch}) and to pinpoint open research challenges with their corresponding research directions (cf. \autoref{sec:challenges}). 
%\janik{The results are presented overall in \autoref{ssec:aggregate} with details per type of requirement in \autoref{ssec:functional} - \autoref{ssec:constraints}. \autoref{ssec:summary} summarizes the assessment results. }

The assessment of the requirements uses an ordinal scale: $+$ means that the requirement is met, $\sim$ means that the requirement is partly met and $-$ means that the requirement is not met. \janikrev{In the following, we illustrate assessment criteria for one selected requirement of each of the $3$ requirement groups (functional, attributes, and constraints). The selection is made due to space limitations (for the remaining requirements, see Section C in the appendix) and aims at conveying the general idea of the assessment criteria. }

\janikrev{
\begin{enumerate}[leftmargin=1.5em,listparindent=0em]
    \item \textbf{Probabilistic Prediction}: 
    %For the probabilistic prediction requirement, 
    $+$ corresponds to directly providing probabilities of occurrence for predicted events. \janikrr{\cite{ning_staple_2018}, for example, propose a multi-task spatio-temporal event forecasting framework to predict the probability of large-scale societal events like civil unrest through news articles. By explicitly predicting the probability $P_i^k$ of the $i$th-societal event occurring in city $k$ given the news articles $X_i^k$ published $H$ days before the event, \cite{ning_staple_2018} meets the requirement.} $\sim$ corresponds to an EP method that does not directly provide occurrence probabilities of predicted events, but can easily be extended to do so. \janikrr{For example, \cite{liu_new_2004} estimate a probability density function for event occurrences across time and space given past event occurrences to assess the crime risk in a monitored region. Evidently, the density function does not constitute an event probability yet, but we can easily compute a probability for a concrete event.} Lastly, $-$ corresponds to a predicted event without an occurrence probability and no easy extension. \janikrr{\cite{cao_remaining_2021}, for example, predict the remaining time that an ongoing process instance takes to complete. Consider a loan application process in which the bank is interested in the remaining times of ongoing applications. For a given application, a LSTM model ensemble is learned from similar past applications. Thus, the approach in \cite{cao_remaining_2021} is not easily extensible to predict the remaining time's probability.}
    \item \textbf{Explainability}: 
    %For the explainability requirement, 
    $+$ corresponds to an approach being ``white box'' (cf. \autoref{sec:arch}).  \janikrr{For example, \cite{yang_web-log_2002} propose a temporal extension to association rules}. $\sim$ corresponds to ``white box'' and ``black box''. \janikrr{\cite{alves_context-aware_2022}, for example, combine transition systems and LSTM to predict remaining times of ongoing process instances}. Lastly, $-$ corresponds to ``black box''. \janikrr{For example, \cite{liu_multi-channel_2022} propose a multi-channel fusion LSTM to predict medical events.}
    \item \textbf{Online Conceptualization}: 
    %For the online conceptualization requirement, 
    $+$ corresponds to an EP method that is defined on an event stream such that the model and predictions are continuously updated and the updating times are known. \janikrr{For example, \cite{kang_carbon_2017} propose EP within the PCM system (cf. \autoref{ssec:arch}) to predict large-scale societal events. Hence, periodic data ingest through Apache Spark Jobs is coupled with daily predictions, a lead time of 5 days (cf. \autoref{ssec:problem}), and a weekly update of all prediction models.} Next, $\sim$ corresponds to an EP method that is defined on an event stream such that the model and predictions are continuously updated, but the updating times are not known. \janikrr{For example, \cite{sahoo_critical_2003} propose to predict system performance and errors for a large computing cluster. Their proposal takes the 5 minute interval for new events and performance measurements into account, but the updating of the prediction models is unknown}. Lastly, $-$ corresponds to an EP method that is defined on a set of historical events with a single training and prediction for evaluation. \janikrr{For example, \cite{sindhu_neuro-genetic_2006} propose to predict security-breaking attacks from past attacks without continuous updates.}
\end{enumerate}}

\subsection{Overall Results}
\label{ssec:aggregate}

\begin{figure}[htb!]
  \centering
  \includegraphics[width=0.85\linewidth]{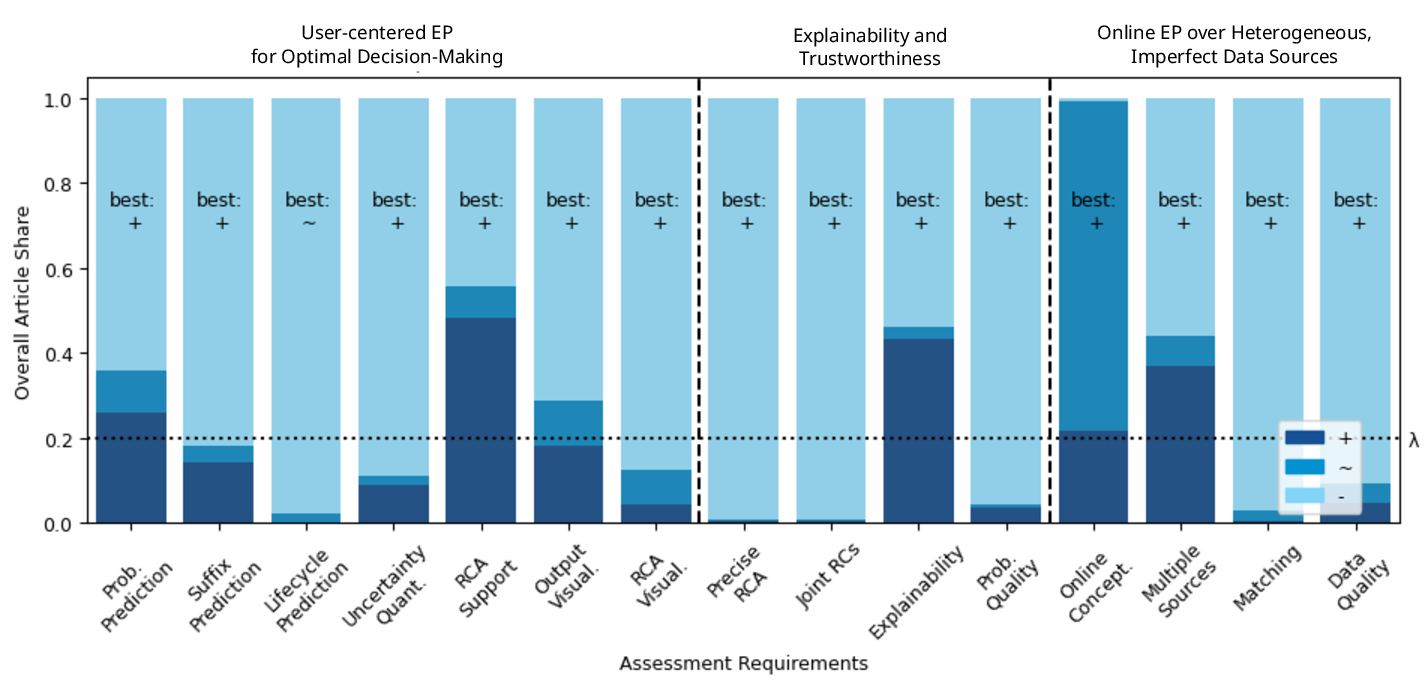}
  \caption{Assessment results for all papers per requirement.}
  \label{fig:sum}
  \Description{Illustration of overall assessment results showing that the respective best source per requirements is mostly meeting the requirements and that there is a lack of sources successfully meeting the requirements for lifecycle prediction and matching.}
\end{figure}

\begin{comment}

\begin{wrapfigure}{R}{4cm}
  \centering
  \includegraphics[width=0.8\linewidth]{boxplot}
  \caption{Distribution of $+$ assessment result per paper.}
  \label{fig:box}
  \Description{Illustration of $+$ assessment result distribution per paper through a boxplot showing that the majority of papers has between one to three $+$ assessed criteria. The lower whisker is at zero and the upper whisker is at six $+$ assessment results. Furthermore, there are two outliers at seven and eight $+$ assessment results.}
\end{wrapfigure}
\end{comment}
\janikrev{Detailed assessment results are reported in our online repository\supplementary.}
The results for each requirement in \autoref{fig:sum} give a nuanced view on research progress in the field of EP methods, \janik{as the overall article shares vary across the requirements. By applying a threshold of $\lambda = 20$ \% (corresponds to 52 articles) on the $\sim$ assessment result, we obtain two sets of requirements: One set that contains requirements with article shares above the threshold (i) and another set that contains requirements below the threshold (ii). In set (i), existing work has proposed methods that meet the requirements \emph{probabilistic prediction}, \emph{RCA support}, \emph{output visualization}, \emph{explainability}, \emph{online conceptualization} and \emph{multiple sources}. The majority of methods} meeting those requirements emphasize the need for monitoring, causality, explanation, and visualization and acknowledge that relevant data is typically distributed over multiple data sources. \janik{In set (ii)}, existing work misses requirements \emph{suffix prediction}, \emph{lifecycle prediction}, \emph{uncertainty quantification}, \emph{RCA visualization}, \emph{precise RCA}, \emph{joint RCs}, \emph{probabilistic quality},  \emph{matching} and \emph{data quality} to a large extent. Despite the very low shares of articles meeting requirements \emph{precise RCA}, \emph{joint RCs}, and \emph{matching}, still some approaches fully meet them ($+$). \emph{Lifecycle prediction} is the only requirement with a $\sim$ as the best assessment.

The distribution of the $+$ assessment count for each paper is presented as Figure \janikrr{3} in the supp. material\supplementary. It reveals that the majority of papers focuses on one to three of the assessment requirements, i.e., the design of EP methods is typically aimed towards a few, particular properties. Although a method design with a clear focus is beneficial during method design, the result leaves the problem of system design to the practitioner. \janikrev{If a paper had the best assessment of meeting all requirements, then the boxplot in the appendix would show a count of 15. As the highest count is an 8, the boxplot shows that the mostly positive best assessment in the requirements view (cf. \autoref{fig:sum}) is due to various papers, i.e., there is no single paper that is close to the best possible assessment.}

\begin{figure}[ht!]
  \centering
  \includegraphics[width=\linewidth]{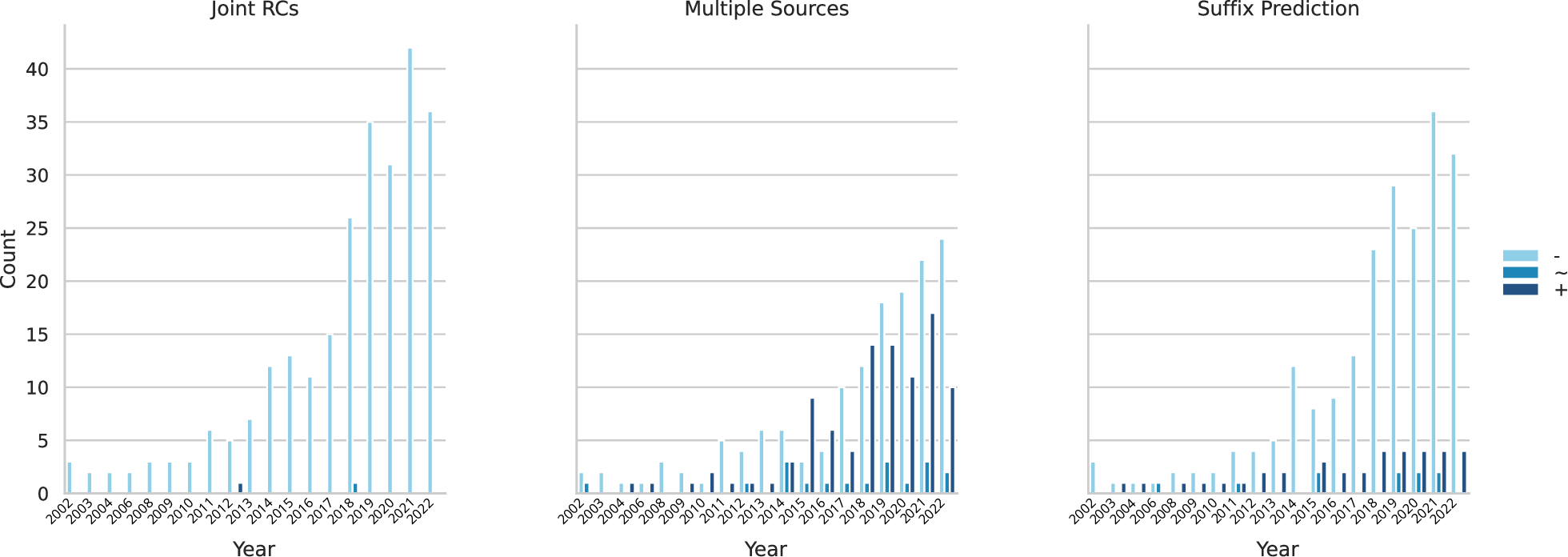}
  \caption{Meeting criteria for requirements \emph{joint RC}, \emph{multiple sources}, and \emph{suffix prediction} (+, $\sim$, -) over time.}
  \label{fig:trends_selected}
  \Description{The meeting criteria trend for \emph{joint RC} shows only very few occurrences of methods that meet the requirement. The meeting criteria trend for \emph{multiple sources} shows a clear upward trend in the number of methods that meet the requirement. The meeting criteria for \emph{suffix prediction} shows a plateau of number of articles at 4 methods per year that meet the requirement.}
  \vspace{-1em}
\end{figure}

\stefanierev{We also systematically analyze the dynamics of EP research based on different timeline-based charts. The distribution of all selected publications over time shows the ongoing and still increasing research output in EP and PPM. We plot the development of the publications regarding their fulfillment of the 15 analyzed requirements per requirement over time. Figure \ref{fig:trends_selected} depicts the plots for requirements \emph{joint RC}, \emph{multiple sources}, and \emph{suffix prediction} as representatives for different patterns/trends.  }

\stefanierev{\emph{Joint RCs} (left side) can be seen as representative distribution for requirements \emph{lifecycle prediction}, \emph{precise RCA}, \emph{prob. quality}, and \emph{matching}. For all of these requirements, $\lambda < 0.2$ and even more precisely, $\lambda \leq 0.04$ holds and the requirement is met (+) or partly met ($\sim$) only selectively. \emph{Multiple sources} (middle part) can be seen as representative distribution for requirements \emph{prob. prediction}, \emph{RCA support}, \emph{output visualization}, \emph{explainability}, and \emph{online concept.}. For all of these requirements, $\lambda > 0.2$ holds and the distribution of criteria met (+), partly met ($\sim$), and not met (-) is evenly distributed with steady increase. %For \emph{prob. prediction}, the distribution of criteria met (+) and partly met ($\sim$) shows steady increase. 
\emph{Suffix prediction} (right part) can be seen as representative distribution for requirements \emph{uncertainty quant.}, \emph{RCA visual.}, and \emph{data quality} for which $0.04 < \lambda < 0.2$ holds. Here, the number of publication with criterion not met (-) increases whereas the other two assessment as + and $\sim$ show a plateau distribution, i.e., the number of publications remains roughly the same over time. }
The distribution of publications categorized by \janikrr{\emph{model type} (cf. Section B in the suppl. material\supplementary)} over the considered timeline reveals two additional interesting observations (cf. Figure \janikrr{4} in the supp. material). First, publications featuring \emph{inferential} prediction models that are ``black box'' follow the overall increasing trend. Interestingly, for \emph{hybrid} models that combine ``black'' and ``white box'', it seems that a peak was reached in 2019, i.e., the number of publications featuring hybrid models has been decreasing since 2019. 

In the following, we aim to support a more holistic development of \janik{EP} methods meeting more requirements by illustrating a \janik{method's properties}\footnote{\janik{Some properties are illustrated with definitions or equations from the original work cited at the beginning of the example, i.e., if the illustration states a definition or equation, then it is taken from the original work and its key part for illustration is extracted and adapted for consistent presentation w.r.t to the survey's notation.}} for each requirement with an example from existing work on each requirement. \janik{The examples are selected based on how straightforward, concise and representative their properties for meeting the requirements are and how well they reflect the diversity of prediction goals and application domains.} The illustration shows how key design aspects are responsible \janik{for properties} of the method that are, \janik{in turn}, responsible for meeting the requirement. %\janik{Please note that we overload the meaning of features as the encoded input data (cf. \autoref{def:prediction}) and the properties of a method that are responsible for meeting a requirement.} 
The presentation of examples is structured along the \janikrr{three themes underlying requirements (cf. \autoref{sec:arch}).} It is accompanied by an analysis of how the requirement may be related to other requirements and a discussion of the respective approaches. 

\subsection{User-centered Event Prediction for Optimal Decision-Making}
\label{ssec:functional}

\janikrr{Seven requirements (cf. \autoref{fig:sum}) center around the user and the support for optimal decision-making given the status of the PCM system (cf. \autoref{sec:arch}). While probabilistic, detailed, and certain predictions across the whole future time horizon are crucial for optimal decision-making, advanced visualization functionalities are insightful to the user.}
 
\begin{figure}[htb!]
  \begin{minipage}{.49\linewidth}
    \centering
    \includegraphics[width=0.25\linewidth]{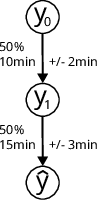}
    \caption
      {%
        An event graph for events $y_0$, $y_1$ and $\hat{y}$ that follow each other in system logs of a supercomputing cluster node \cite{fullop_real_2012}. The probability of following, the average time delay and the standard deviation of time delay is added for each relation between events.%
        \label{fig:event_graph}%
      }%
      \Description{Illustration of an event graph with three events $y_0$, $y_1$ and $\hat{y}$ and the respective probability of following, average time delay and the standard deviation of time delay \cite{fullop_real_2012}. The probability of following between $y_0$ and $y_1$ is 50\% and the average time delay is 10min with standard deviation +/- 2min. The probability of following between $y_1$ and $\hat{y}$ is 50\% and the average time delay is 15min with standard deviation +/- 3min.}
  \end{minipage}\hfill
    \begin{minipage}{.49\linewidth}
    \centering
    \includegraphics[width=0.9\linewidth]{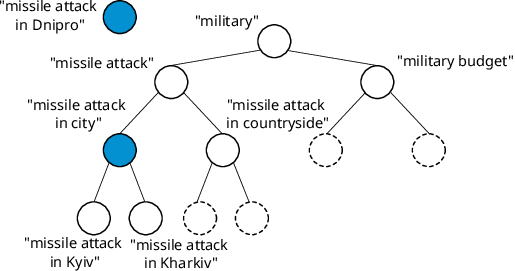}
    \caption
      {%
        A hierarchy of most general to most specific observed events \cite{radinsky_learning_2012}. On receiving a current event of a missile attack in Dnipro, the most specific event in the hierarchy that semantically matches the current event is used for prediction. %
        \label{fig:jointrc}%
      }%
      \Description{Illustration of a semantic hierarchy of observed events \cite{radinsky_learning_2012}. The most general observed event pertains to the "military". Below "military", there are "missile attack" and "military budget". "Missile attack" is further specialized into "missile attack in city" and "missile attack in countryside" of which "missile attack in city" is further specialized into "missile attack in Kyiv" and "missile attack in Kharkiv". The current event is "missile attack in Dnipro" which is matched with "missile attack in city".}
  \end{minipage}\hfill

\end{figure}

\textbf{Probabilistic prediction.} \cite{fullop_real_2012} predicts the time and event failure type of high-performance computing cluster nodes by extracting event graphs from system logs. \janikrr{To that end, \cite{fullop_real_2012}} compute the confidence of the correlation between events, the average time delay between events and its standard deviation. \autoref{fig:event_graph} shows a small, exemplary event graph with three events $y_0$, $y_1$ and $\hat{y}$. For a \emph{probabilistic prediction}, \janik{the EP method in \cite{fullop_real_2012} predicts the probability of event $\hat{y}$ that follows event $y_0$ and $y_1$ with Bayes law to be 50\% * 50\% = 25\% and the time of event $\hat{y}$ to be 25 minutes +/- (2 * 4 minute + 2 * 3 minute) = 14 minutes with 95\% probability after $y_0$ has occurred.}

Leveraging the event graph, Bayes law and the assumption of normally distributed time delays leads to a straightforward, probabilistic conceptualization of EP. For the same reasons, the prediction can be traced back to one or a set of root causes in the event graph such that the method meets the requirements of \emph{RCA support}. Furthermore, the simplicity of the method helps domain experts to understand its output. \janikrr{Yet,} it may fail to perform well when facing complex dependencies between events in multiple dimensions, \janik{e.g., a patient's clinical history and future treatment outcome. This indicates a trade-off between simplicity of the probabilistic conceptualization and its applicability to complex monitored systems, their application domains and corresponding prediction goals.}

Methods with a probabilistic prediction all conceptualize the method with conditional probabilities, Bayesian or Markovian settings, but differ in their degree of statistical rigor. Statistical rigor captures the number of assumptions on the distribution of the input data, how restraining the distribution assumptions are and how formal the derivation of the EP method is given those assumptions. For example, \cite{antunes_bayesian_2003} exhibits a high degree of statistical rigor, as it features a high number of assumptions on the distribution of the data that restrain it to a large degree (e.g., the input data is the result of a Gaussian autoregressive process of the second order). \janikrr{Then,} the whole EP method is formally derived from those assumptions. \cite{fullop_real_2012} exhibit a medium degree of statistical rigor, as they assume normally distributed time delays and apply Bayes law, but do not formally derive the overall EP method. \cite{pickett_random_2021} exhibits a low degree of statistical rigor, as they explicitly aim at designing the EP method to have no assumptions and do not formally derive the method. 

\textbf{Suffix prediction.} \cite{kang_periodic_2012} predict the full sequence of future events for an ongoing manufacturing cycle. \janikrr{Thus, \cite{kang_periodic_2012} predicts} all the events that are likely to occur until the production is complete. \janikrr{The prediction goal is a set of key performance indicators such as product quality}. \cite{kang_periodic_2012} assume that the future progress of the ongoing cycle $s$ with length $n$ can be similar to the progress of $k$ historical cycles $l_1, \ldots, l_k$ after $n$ events of each historical cycle $l_i$ have been observed. Based on this assumption,  \cite{kang_periodic_2012} apply k-Nearest-Neighbor clustering with Euclidean distance on the $n$-length prefix of historical cycles and the ongoing cycle resulting in $k$ suffixes for the ongoing trace. Then,  \cite{kang_periodic_2012} apply a support vector machine classifier on the $k$ suffixes to predict the key performance indicator. At last, the $k$ predicted key performance indicators are aggregated into a single indicator. 
Although this indirect basic technique can show the operations manager in the production facility additional information on how the current cycle will progress, this method likely fails to accurately predict for a production line with hundreds of thousands of different cycle variations. In contrast to \cite{kang_periodic_2012}, many other methods such as \cite{lin_mm-pred_2019} or \cite{di_francescomarino_eye_2017} predict the suffix by repeatedly applying a next EP method. Although this can be done for any EP method, the major challenge of looping through events that are likely has to be tackled by methods repeatedly applying the EP method for \emph{suffix prediction}.

\textbf{Lifecycle prediction.} \cite{lin_mm-pred_2019} predict the next events of business processes by learning a recurrent neural network (RNN) that takes multiple event attributes into account. Accordingly, the EP problem is stated for an ongoing process instance $y = \langle (e_1, a^1_1, \ldots , a^m_1), \ldots , (e_n, a^1_n, \ldots , a^m_n) \rangle $ with event activities $e_1, \ldots, e_n $ and event attributes $a^1_1, \ldots , a^m_n $ as predicting the next event $(e_{n + 1}, a^1_{n + 1}, \ldots , a^m_{n + 1})$ of the ongoing process instance $s$. Hence, \cite{lin_mm-pred_2019} can predict lifecycles of events, since the lifecycle of an event is an attribute of the event. Nevertheless, lifecycle carries special semantics for EP. \janikrr{For example,} an \emph{abort} lifecycle event for the delivery activity of a logistics company signifies that the \emph{end} will not occur at all and, therefore, the delivery will fail \cite{rinderle-ma_predictive_2022}. Also, the lifecycle determines the duration of the activity recorded through multiple events. \janikrr{For instance,} an activity paused once for 5 minutes has a duration of 5 minutes lower than the time between its \emph{start} and \emph{end}. For prediction goals that are evaluated on the duration, e.g., efficiency of the resource executing the activity, the special semantics of the lifecycle has to be taken into account. Consequently, the prediction method should not stop at predicting the lifecycle as an attribute, but incorporate its semantics. 

Similarly, the other proposals partly meeting lifecycle prediction conceptualize EP with multiple, unspecified event attributes and do not incorporate lifecycle semantics. If the data source for the evaluation does not provide lifecycle information, this should not be taken as the reason for excluding it in a method design. \janikrr{Exclusion} would not allow the method to be generally applied to data sources with event lifecycle information. Due to the importance of lifecycle information for prediction, researchers should recommend the additional recording of this information to a data source, if the data source does not provide lifecycle information.

\textbf{Uncertainty quantification.} \cite{wu_uncertainty-aware_2021} predict the time-to-event for treatment outcomes of patients in clinics by extracting features with RNNs and learning a Parametric Predictive Gaussian Process (PPGP) Regressor in a survival analysis setting. The combination of the PPGP Regressor and survival analysis results in the survival function:
\[
    S(y_i | \mathbf{h}_i) = 1 - F\left(\dfrac{y_i - \mu_\mathbf{f}(\mathbf{h}_i)}{\sigma_\mathbf{f}(\mathbf{h}_i) + \sigma_\text{obs}}\right) 
\]
where $S$ is a survival function, $y_i$ the logarithm of the time-to-event for patient $i$, $\mathbf{h}_i$ the latent representation extracted by the recurrent neural network, $ F$ the normal cumulative distribution function, $\mathbf{f} \in \mathbb{R}^n$ a vector of Gaussian process function values, $\sigma^2_\mathbf{f}(\mathbf{h}_i)$ the input-dependent predictive variance and $\sigma_\text{obs}^2$ the input-independent observational noise for the logarithmic time-to-event $y_i$. The corresponding objective that \cite{wu_uncertainty-aware_2021} propose puts more weight on the input-dependent predictive variance $\sigma^2_\mathbf{f}(\mathbf{h}_i)$ than earlier proposals that combine Gaussian Processes and survival analysis. Hence, the formulation of survival analysis with PPGP Regressors and the corresponding uncertainty-aware objective quantifies predictive variance, i.e., epistemic uncertainty%\footnote{Model uncertainty does not exist due to an implicit assumption that the true predictor is part of the chosen model hypothesis space \cite{hullermeier_aleatoric_2021}.}
, and observational noise, i.e., aleatoric uncertainty. \janikrr{As well,} the formulation also strengthens the positive relationship between predictive variance and prediction error. Consequently, the less confident the uncertainty-aware EP method is, the less accurate its predictions are.

In addition to quantifying the uncertainty of the prediction, appropriately integrating it to the objective of the EP method yields a desirable relationship between uncertainty and prediction quality. The information that more confident predictions correspond to more accurate predictions is valuable to practitioners \cite{hullermeier_aleatoric_2021}. \janik{This information} likely results in more trust towards the method. By further differentiating between epistemic and aleatoric uncertainty, the quantified uncertainty also shows whether a currently low confidence could be improved with more data.

Existing proposals mostly quantify uncertainty by means of a probabilistic conceptualization that allows to inherently estimate uncertainty as some form of variance in the method and, thus, focuses on the method, e.g. \cite{wu_uncertainty-aware_2021}. Depending on the degree of statistical rigor (cf. \emph{probabilistic prediction}), these estimates are more or less justified by statistical theory. Another option is to extrinsically quantify uncertainty by approximating it using, e.g. characteristics of the input data. \cite{comuzzi_does_2019} propose an uncertainty metric that also considers how many activities of a business process instance have been observed and what time has elapsed since the last activity has been observed. The rationale for their extrinsic uncertainty quantification is that more and more recent data on the monitored system relate to a more certain prediction.

\textbf{RCA support, output visualization and RCA visualization.} While \cite{fullop_real_2012} apply Bayes law on descriptive statistics of the data without a statistical analysis formulation, \cite{letham_interpretable_2015} combine decision rules with Bayesian analysis into Bayesian Rule Lists (BRL): 

{\centering
  $ \displaystyle
    \begin{aligned} 
    &\text{\textbf{if}} \, \text{antecedent}_1 \, \text{\textbf{then}} \; y \sim \text{Multinomial}(\mathbf{\Theta}_1), \, \mathbf{\Theta}_1 \sim \text{Dirichlet}(\mathbf{\Phi} + \text{\textbf{N}}_1) \\
    &\vdots \\
    &\text{\textbf{else if}} \, \text{antecedent}_m \, \text{\textbf{then}} \; y \sim \text{Multinomial}(\mathbf{\Theta}_m), \, \mathbf{\Theta}_m \sim \text{Dirichlet}(\mathbf{\Phi} + \text{\textbf{N}}_m) \\
    &\text{\textbf{else}} \, y \sim \text{Multinomial}(\mathbf{\Theta}_0), \, \mathbf{\Theta}_0 \sim \text{Dirichlet}(\mathbf{\Phi} + \text{\textbf{N}}_0) \\
\end{aligned}
  $ 
\par}

\noindent where antecedents are well-formed formulas with $c_i, \, i \in \{1, \ldots, m\}$ atomic propositions over the feature vector that do not overlap, $y$ is the prediction goal value, $\Theta_i$ are parameters of the respective distribution, $\Theta_0$ a default rule parameter for observations that do not satisfy any of the antecedents, $\mathbf{\Phi}$ a prior of the prediction goal values and $\text{\textbf{N}}_i$ the respective observation counts satisfying antecedent $i$. By limiting the parameters $m$ and $c_i$ to small integers as well as deriving a mean point estimate and credible interval for the prediction goal value $y$, BRLs represent a simple, interpretable prediction model. Potential root causes are represented as antecedents of the BRL (\emph{RCA support}). The prediction output and the RCA are visualized by showing the BRL and traversing its rules from top to bottom given an observation. 
\begin{comment}

The following BRL predicts the stroke risk of atrial fibrillation patients\todo{Figure as text}: 
\todo{Potentially cut this for space}
{\centering
  $ \displaystyle
    \begin{aligned} 
\text{"}&\text{\textbf{if} hemiplegia and age} > 60 \, \text{\textbf{then} stroke risk} \; 58.9\% \, (53.8\% \,\text{-}\, 63.8\%) \\
&\text{\textbf{else if} cerebrovascular disorder \textbf{then} stroke risk} \; 47.8\% \, (44.8\% \,\text{-}\, 50.7\%) \\
&\text{\textbf{else if} transient ischaemic attack \textbf{then} stroke risk} \; 23.8\% \, (19.5\% \,\text{-}\, 28.4\%) \\
&\text{\textbf{else if} occlusion and stenosis of carotid artery without infarction \textbf{then} stroke risk} \; 15.8\% \, (12.2\% \,\text{-}\, 19.6\%) \\
&\text{\textbf{else if} altered state of consciousness and age} > 60 \, \text{\textbf{then} stroke risk} \; 16.0\% \, (12.2\% \,\text{-}\, 20.2\%) \\
&\text{\textbf{else if} age} \leq 70 \, \text{\textbf{then} stroke risk} \; 4.6\% \, (3.9\% \,\text{-}\, 5.4\%) \\
&\text{\textbf{else} stroke risk} \; 8.7\% \, (7.9\% \,\text{-}\, 9.6\%)\text{"} \;\text{\cite{letham_interpretable_2015}}\\
\end{aligned}
  $ 
\par}
\end{comment}
\cite{letham_interpretable_2015} compares this BRL to a medical score that is frequently used to predict the stroke risk by physicians. \janikrr{They} identify antecedents with known root causes or risk factors and conclude that the BRL is as interpretable as the medical score, while improving the accuracy. 

Methods meeting all three requirements \emph{RCA support}, \emph{output visualization}, and \emph{RCA visualization} are designed in their functionality to support the user in understanding the method and its prediction. A simple prediction model as in \cite{fullop_real_2012,letham_interpretable_2015} underlying the EP method support the user in tracing the prediction back to its root causes. This meaning of simplicity is captured by the "white box" assessment of \emph{explainability}, but it is not the only sufficient condition for meeting \emph{RCA support}. \janikrr{For instance,} a hybrid basic approach can still support RCA in spite of a "black box" model due to the ``white box'' part of the model \cite{lindemann_anomaly_2020}. \janik{Due to a lack of a widely accepted definition on \emph{explainability} (cf. \autoref{sec:arch}), it is not clear whether further relationships with root causes exist}. Considering visualization, some methods present a straightforward way of visualizing the output or RCA as in \cite{letham_interpretable_2015}, while others need specifically designed visualization techniques, e.g. neural networks. 

\subsection{Explainability and Trustworthiness}
\label{ssec:attributes}
\janikrr{Four requirements (cf. \autoref{fig:sum}) aim towards explainable and trustworthy EP through appropriate quality metrics and root-cause analysis}.

\textbf{Precise RCA, joint RCs and explainability.} \cite{radinsky_learning_2012} predict future events that can be caused by a current news event. To that end, they choose an aggregate semantic event representation for an event $e = (P, O_1, \ldots , O_4, t)$. Here, $P$ is a state or temporal action exhibited by the event's objects, $O_1$ a set of users, $O_2$ a set of objects on which the action $P$ was performed, $O_3$ a set of instruments utilized by the action, $O_4$ a set of locations and $T$ a timestamp. Given a set of historic events  $\{\langle e_1, g(e_1) \rangle, \ldots , \langle e_n , g(e_n) \rangle \}$ for an unknown causality function $g$ mapping a historic cause event to its effects, \cite{radinsky_learning_2012} aim to approximate $g$ by learning a causality graph of events. \janikrr{The causality graph is learned through} existing ontologies on actions such as VerbNet \cite{kipper2006extending} and on objects such as the LinkedData ontology \cite{bizer2011linked}. 

Using existing ontologies, the method can evaluate the precision of identified root causes (\emph{precise RCA}). For identifying \emph{joint RCs}, the method proposes a generalization event by generalizing over its actions and objects. The result is a hierarchy that detects that similar root causes are in fact joint root causes. \autoref{fig:jointrc}, for example, depicts a generalization hierarchy of observed events that identifies similar root causes as joint, more general root causes. The prediction is computed solely on the causality graph through matching the current event with an existing event in the causality graph (denoted in green in \autoref{fig:jointrc}) and applying a prediction rule associated with the event. \janikrr{Thus,} a user can traverse the predicted event back to its historic event. Finally, the method employs a ``white box'' model (\emph{explainability}).  

Despite their existence, ontologies are rarely used in EP. With the help of ontologies, the method can establish a concept of causality and meaning. Moreover, EP methods rarely exploit that events are related by a semantic generalization relationship. However, other semantic relationships between events are exploited by recent methods, e.g. the disease co-occurrence \cite{lu_context-aware_2022} or compatible related treatment outcomes relationships \cite{gao_clinical_2022}. 

\textbf{Probabilistic Quality.} \cite{wang_csan_2020} predict crime frequencies in the spatio-temporal domain by combining variational autoencoders and sequence generative neural networks. Hence, a predicted crime frequency is $\hat{y}_{i,j,k}^{(t)}$ for the geographical grid indexes $i, j$, crime type $k$ in the time slot $t$. To evaluate these probabilistic predictions, \cite{wang_csan_2020} first normalize the true and predicted crime frequencies: 

\[
    p_{i,j,k}^{(t)} = \dfrac{y_{i,j,k}^{(t)}}{\sum_{i=1}^{M}\sum_{j=1}^N y_{i,j,k}^{(t)}}  , \quad   \hat{p}_{i,j,k}^{(t)} = \dfrac{\hat{y}_{i,j,k}^{(t)}}{\sum_{i=1}^{M}\sum_{j=1}^N \hat{y}_{i,j,k}^{(t)}}
\]
and then compute the Jensen-Shannon divergence with laplacian smoothing:

\[
    D_{\text{JS}}(P, \hat{P}) = \dfrac{1}{2}D_{\text{KL}}(P \,\|\,Q) + \dfrac{1}{2}D_{\text{KL}}(\hat{P} \,\|\, Q)
\]
where $Q = \dfrac{1}{2} (P+\hat{P})$ is the average of true and predicted probability. Next to the Jensen-Shannon divergence, there exist at least 200 further scores for evaluating probability estimates \cite{doi:10.1287/deca.2016.0337}. Thus, analysing its properties with respect to EP and proposing a standardized set of scores goes beyond the scope of this survey. For this survey, the important point is that the evaluation is done using a metric that scores probability estimates, as is the case for \cite{wang_csan_2020}.

For evaluating probabilistic predictions, the predictions must come with probability estimates. Thus, meeting the requirement of \emph{probabilistic quality} implies meeting \emph{probabilistic prediction}. Yet, these two requirements are not equivalent, because many articles do not report results of scoring probability estimates. While roughly 25\% of the 260 articles propose methods meeting \emph{probabilistic prediction}, less than 5\% evaluate the predicted probabilities with an appropriate score (cf. \autoref{fig:sum}). 

\subsection{Online Event Prediction over Heterogeneous, Imperfect Data Sources}
\label{ssec:constraints}

\janikrr{Four requirements (cf. \autoref{fig:sum}) represent the need for continuous EP given heterogeneous data sources with dynamically-changing data quality}.

\textbf{Online conceptualization.} \cite{ramakrishnan_beating_2014} predict civil unrest events across ten Latin American countries through a continuous, automated system. \janikrr{The system} runs 24/7 and constantly processes events coming from various open data sources such as social media. Consequently, the EP method consisting of an ensemble of methods is proposed to continuously learn from daily data. \janikrr{For example}, a logistic regression models is learned using daily tweet counts while a rule-based method detects keywords contained in social media posts. The system has a throughput of 200-2000 messages/sec, predicts roughly 40 events/day and can ingest up to 15 GB of messages on a given day. Therefore, it features an \emph{online conceptualization} and can monitor the ten Latin American countries for social unrest events. 
Given its daily cycle and the ensemble of prediction methods, a daily, full retraining of the prediction models is feasible. However, a full retraining may become infeasible \janik{for EP methods in light of updating cycles not longer than a few minutes}. For this reason, \cite{rizzi_how_2022} investigate strategies for updating the prediction model that go beyond full retraining: \texttt{Do nothing}, i.e., no update; \texttt{retraining with no hyperparameter optimization}, i.e., a lightweight retraining; \texttt{full retraining}, i.e., train on all available events; and \texttt{incremental update}, i.e., applying incremental learning algorithms. The results show that \texttt{full retrain} and \texttt{incremental update} are the best strategies for predicting business process events. 

Still, not all prediction models allow for incremental updates through incremental learning algorithms. Thus, \cite{marquez-chamorro_updating_2022} investigates six strategies to handle updates to the prediction model by means of data selection strategies. \janikrr{First,} the \texttt{baseline} \janikrr{strategy means} no update to the training data. \janikrr{Second,} the \texttt{cumulative} \janikrr{strategy means} ``update training data on every new event''. \janikrr{Third,} the \texttt{non-cumulative} \janikrr{strategy means} ``update training data on every new event by keeping the $k$ most recent events''. The \texttt{ensemble} \janikrr{strategy means} ``non-cumulative and keeping all models for ensemble prediction''. \janikrr{Next,} the \texttt{sampling} \janikrr{strategy means} ``update training data by sampling all available events''. \janikrr{Finally,} the \texttt{drift} strategy is non-cumulative with drift detection such that only training data after a drift is selected. The results show that the \texttt{ensemble} strategy performs best for predicting business process events.

\janik{Despite these pointers to suitable strategies for updating prediction models, a best strategy for all use cases cannot be determined due to a case-dependent trade-off between the resources required to improve the quality of prediction through updating and its benefits.}

\textbf{Multiple sources.} \cite{aggarwal_two_2018} predict the time-to-event for future failures of engineering system devices. Sensor data  $X_p \in \mathbb{R}^{d \times c_p}$ from $d$ sensors for each device $p$ observed until time $c_p$ and event data $y_p \in \{0, 1\}$ is used to predict the time-to-event $T$ for the future failure $y_p = 1$ of device $p$. For prediction, \cite{aggarwal_two_2018} propose a multi-task learning framework based on neural networks with a task $p$ for each device. Hence, $p$ data sources are used in the EP method (\emph{multiple sources}) and the integration of these data sources is implicitly learned by the neural network. 

\janik{We can distinguish two approaches to data integration. The \emph{implicit} approach to data integration combines the data integration with the learning of a prediction model in the EP method such that the data integration is tightly coupled with the EP method.} \janikrr{For example}, the data integration of the $p$ devices is implicitly learned by the neural network in \cite{aggarwal_two_2018}. The \emph{explicit} approach to data integration separates data integration from the EP method such that both of them become interchangeable. To that end, the explicit approach implements necessary functionality for data integration (cf. input component in \autoref{ssec:arch}) before the EP method is applied to the data. \cite{ramakrishnan_beating_2014} not only illustrates the \emph{online conceptualization}, but also the explicit approach to data integration. Many different sources such as NASA satellite meteorological data, Bloomberg financial news and Twitter's public API are used. Each source is ingested by a specialized routine to convert the data input to JSON and add identifiers. Before the EP method is applied, the JSON input is enriched through, e.g. geocoding, data normalization and entity extraction. Consequently, the EP method does not need to learn how it can integrate the data. In comparison, the implicit approach intertwines the input component of a PCM system (cf. \autoref{ssec:arch}) with the prediction component, whereas the explicit approach keeps them separated. The former has the advantage, that it does not need to design data integration functionalities, but may experience worse performance for heterogeneous data sources and consuming more resources for training. The latter has the advantage, that it has direct control and knowledge on the specific data integration routines, but comes at the cost of more design and maintenance effort. 

\begin{figure}[htb!]
\begin{minipage}{.49\linewidth}
    \centering
    \includegraphics[width=\linewidth]{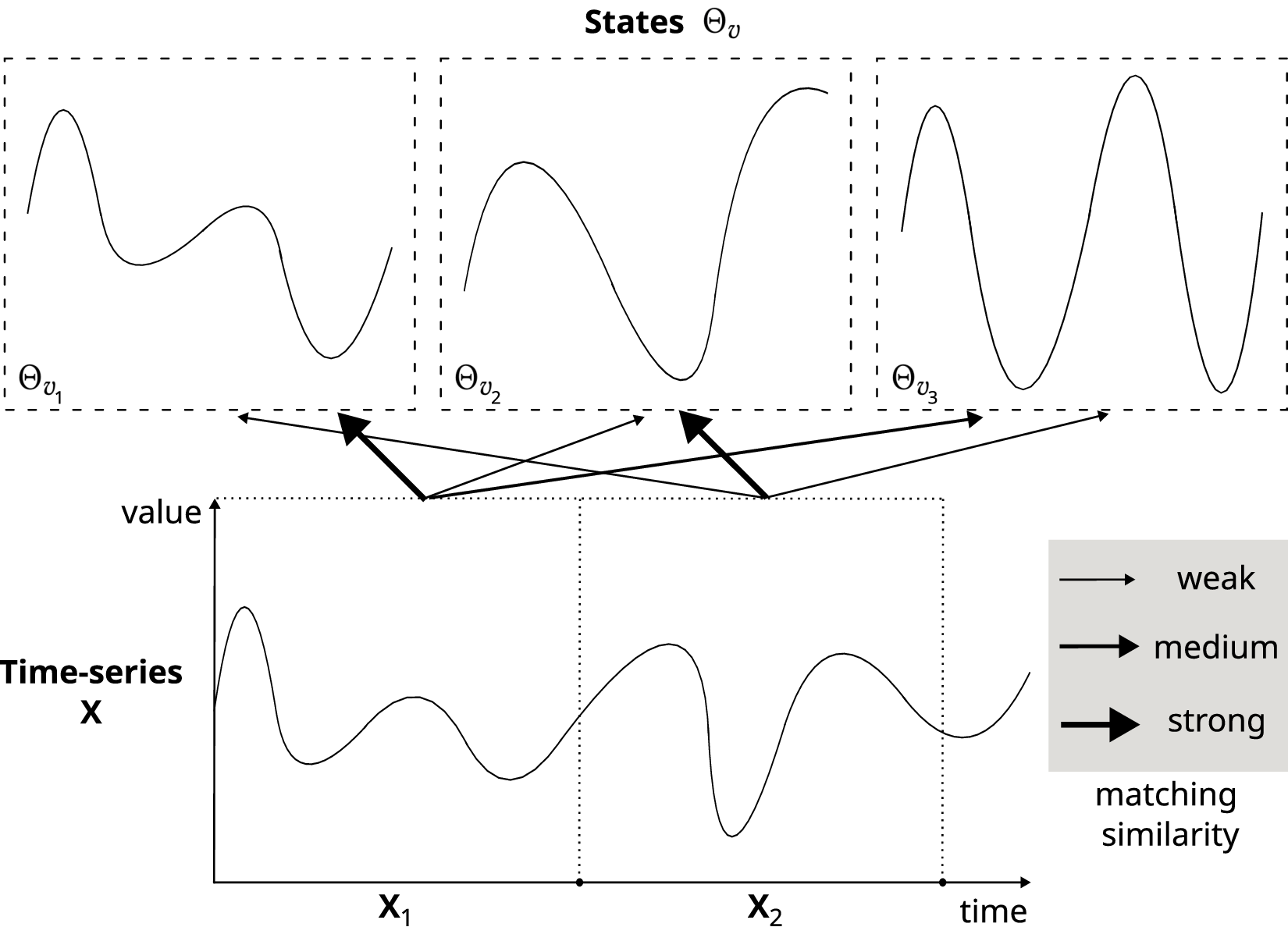}
    \caption
      {%
        Time-series $\mathbf{X}$ is segmented into $\mathbf{X}_1$ and $\mathbf{X}_2$. Each time-series segment is recognized with states $\Theta_{v_1}$, $\Theta_{v_2}$ and $\Theta_{v_3}$ \cite{hu_time-series_2021}. Recognized states and the similarities are used in the EP method proposed in \cite{hu_time-series_2021} instead of the original time-series. %
        \label{fig:evo}%
      }%
      \Description{Illustration of segmented time-series $\mathbf{X}$ into $\mathbf{X}_1$ and $\mathbf{X}_2$. Each time-series segment is recognized with states $\Theta_{v_1}$, $\Theta_{v_2}$ and $\Theta_{v_3}$ \cite{hu_time-series_2021}. }
  \end{minipage}\hfill
  \begin{minipage}{.49\linewidth}
    \centering
    \includegraphics[width=\linewidth]{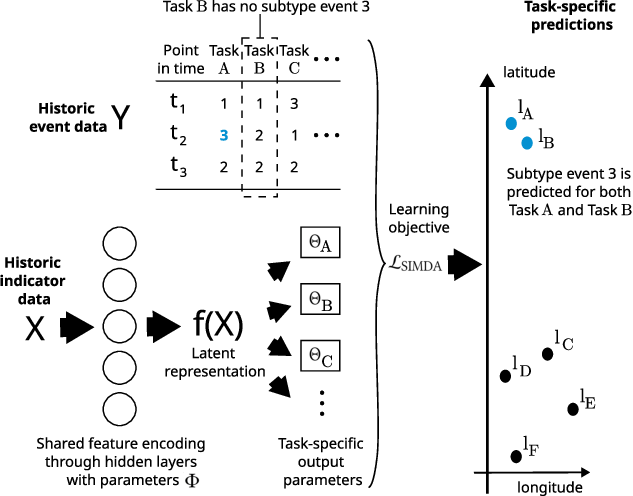}
    \caption
      {%
        Spatial Incomplete Multi-Task Deep LeArning (SIMDA) Framework \cite{gao_incomplete_2019}. Despite no recorded subtype event 3 in the historic event data $Y$ for task $B$, SIMDA allows to predict subtype event 3 in task $B$ that corresponds to location $l_B$, since location $l_A$ is geographically close to location $l_A$ and location $l_A$ has recorded subtype event 3 in the historic event data of task $A$. %
        \label{fig:simda}%
      }%
      \Description{Illustration of Spatial Incomplete Multi-Task Deep LeArning (SIMDA) Framework \cite{gao_incomplete_2019}. SIMDA takes historic event data $Y$ and historic indicator $X$ as input. In the historic event data $Y$, subtype 3 is missing for task $B$ corresponding to location $l_B$, while the historic event data $Y$ shows for the geographically close location $l_A$ corresponding to task $A$ that subtype event 4 occurred. Together with a latent representation of the historic event data $X$ through a shared hidden layer encoding with parameters $\Theta$, the learning objective $\mathcal{L}_\text{SIMDA}$ enforces to have geographically close locations to have more similar event subtype patterns. The result is the ability to predict subtype 3 for location $l_B$ in task $B$, although it has no event subtype 3 in its historic event data.}
  \end{minipage}\hfill
\end{figure}

\textbf{Matching.} \cite{hu_time-series_2021} predict a future event in a time-series sequence, e.g. a future anomalous watt-hour meter clock event of the State Grid of China in weekly sensor readings. To that end, a future event $y_{t + 1} \in \mathbb{Z}$ is predicted based on historic events $y_t$ and $T$ segments $\mathbf{X}_t \in \mathbb{R}^{\tau \times d}$ of the time-series sequence $\mathbf{X} \in \mathbb{R}^{N \times d}$ with length $\tau$: 

\[
    \langle \mathbf{X}_{1:T}, y_{1:T} \rangle = \{ (\mathbf{X}_1, y_1), \ldots , (\mathbf{X}_T , y_T) \} 
\]

Before feeding the historic data $\langle \mathbf{X}_{1:T}, y_{1:T} \rangle$ into a neural network, the proposed EP method recognizes representative time-series patterns called \emph{states} and denoted as $\Theta_v \in \mathbb{R}^{\tau \times d}$ for segments of the data (cf. \autoref{fig:evo}). Hence, redundant or very similar time-series patterns from multiples sources are recognized by the same state (\emph{matching}). 
\cite{hu_time-series_2021} is the only method proposed that features an explicit matching mechanism for data coming from multiple sources. Other proposals for matching \cite{li_constructing_2018,lei_event_2019,zhang_analogous_2020,wang_multi-level_2021,hoegh_bayesian_2015,ramakrishnan_beating_2014,kang_carbon_2017,acharya_causal_2017} are not explicitly including the mechanism in their method. 

\textbf{Data quality.} \cite{gao_incomplete_2019} predict the location of event subtypes. \janikrr{Therefore,} instead of predicting the air pollution of cities as future events, the respective air pollutant subtype (Carbon monoxide, nitrogen dioxide, fine dust pollution etc.) is predicted. The predicted event subtype allows for more fine-grained actions, e.g. allocating appropriate resources per air pollutant subtype. \janikrr{However,} incomplete, historical data for event subtypes in many locations becomes a major data quality challenge for predicting event subtypes for all locations. 

\autoref{fig:simda} depicts the data quality challenge for predicting event subtypes and the proposed method, Spatial Multi-task Deep leArning (SIMDA), to overcome the challenge. Each depicted prediction task $A, B$ and $C$ corresponds to a geographical location $L = \{l_A, l_B, l_C, \ldots \}$. \janik{Historic event data $Y_{l,t} \in S = \{ 1, 2, 3, \ldots \}$ and }$d$-dimensional, historic indicator data $X_{l,t} \in \mathbb{R}^{1 \times d}$ for date $t$ is available for each location $l \in L$. However, for the location $l_B$ corresponding to task $B$, no event subtype $s = 3 $ was recorded, e.g. a city in the past has not recorded fine dust pollution (cf. \autoref{fig:simda}). Hence, the specific model for task $B$ will not predict event subtype $\hat{y}_{l_B, \tau} = 3$ at location $l_B$ for $\tau = t + p, p > 0$, although the geographically close location $l_A$ in task $A$ shows that event subtype $s = 3$ has occurred. Since geographical heterogeneity across all locations does not allow for a single task approach, learning from other locations such as $l_A$ for location $l_B$ has to be achieved differently. \janikrr{Here, the} first law of geography stating that geographically closer locations will be more similar to each other than farther locations \cite{cressie2015statistics} is exploited. \cite{gao_incomplete_2019} state that two close locations $i, j \in L$ have similar conditional probabilities for an event subtype $s$:

\[
    P(Y_{i,t} = s | X_{i, t}) \approx P(Y_{j,t} = s | X_{j, t})
\]

This relationship is exploited to formulate a SIMDA objective that enforces event subtype patterns to be similar for geographically close tasks: 
\[
     \mathcal{L}_\text{SIMDA} = \mathcal{L}_d(\Phi, \Theta) + \dfrac{\beta}{2} \sum_l^L\sum_{s,r}^{S^2}\|f(X_l)(\Theta_{l,s} - \Theta_{l,r})^T - \dfrac{1}{N_l}\sum_c^L adj(l,c)f(X_c)(\Theta_{c,s} - \Theta_{c,r})^T\|_2^2
\]
where $\mathcal{L}_d(\Phi, \Theta)$ is the general multi-task deep learning objective with $d$ dimensions, $\Phi$ the weight parameters of the shared hidden layer, $\Theta$ the weights of the task-specific output layer, $\beta$ a hyperparameter, $f(\cdot)$ the computation of the shared hidden layers, $\Theta_{l,s}$ the weights of task $l$ for predicting event subtype $s$, $N_l = \sum_c^L adj(l,c)$ the normalization term for location $l$, and $adj(\cdot, \cdot)$ some physical distance function. Hence, \cite{gao_incomplete_2019} propose to add a regularization term to the general multi-task deep learning objective $\mathcal{L}_d(\Phi, \Theta)$ that enforces geographically close locations $i, j $ to have similar event subtype patterns $X_i (\Theta_{i,s} - \Theta_{i,r})^T \approx X_j (\Theta_{j,s} - \Theta_{j,r})^T$. For the example as depicted in \autoref{fig:simda}, this allows SIMDA to predict fine dust pollution $s = 3$ for location $l_B$ through task $B$, as the geographically close location $l_A$ exhibits a lot of fine dust pollution.

Analogous to \cite{gao_incomplete_2019}, methods that actively utilize data quality issues exploit the following semantic similarity relationships between components of the monitored system (e.g., close locations, comparable patients etc.): Semantic similarity based on proximity \cite{zhao_hierarchical_2016,zhao_multi-resolution_2016,gao_incomplete_2018,gao_incomplete_2019}, hierarchy of geographical regions \cite{zhao_hierarchical_2016,zhao_multi-resolution_2016}, logging granularity of cyber system host sources \cite{wu_joint_2022}, patient disease histories \cite{soleimani_scalable_2018,pickett_random_2021,lu_self-supervised_2021}, hierarchy of medical codes \cite{lu_self-supervised_2021}, hierarchy of mobile health study participants \cite{dempsey_isurvive_2017}, individual units of engineering systems \cite{deep_event_2020}, feature missingness patterns \cite{wang_incomplete_2018} and structural tweet content \cite{zhao_spatiotemporal_2015}. \janikrr{Exploiting semantic similarity relationships aims to improve the prediction while keeping the component-level granularity in the prediction model (e.g. a task for each location).} In contrast to preprocessing techniques such as imputation for missingness, these proposals make the method aware of the data quality and base the awareness on the semantic similiarity relationships. 

Since data quality issues should lead to more uncertain predictions, this requirement is related to \emph{uncertainty quantification}. For example, \cite{soleimani_scalable_2018} propose a method that abstains from predicting in case of high uncertainty of the prediction due to missingness of data. Hence, the missingness of data has a direct impact on the prediction in their proposed method. All in all, the link between data quality and uncertainty can be used to guide the prediction (e.g., no prediction in case of high uncertainty due to data quality issues) or the data quality issue can be alleviated by exploiting semantic relationships leading to more certain predictions. 

%\subsection{Trends and Patterns}
%\label{ssec:trends}

\subsection{Summary}
\label{ssec:summary}

%We can conclude from the overall results in \autoref{ssec:aggregate} and the results for the three types of requirements in \autoref{ssec:functional} - \autoref{ssec:constraints} that research on EP as a whole shows considerable progress in almost every of the 16 requirements. The individual contributions are varied and each highlight how a combination of the prediction goal and the EP method are used to achieve the respective requirement. Although we have at least a solution to each requirement except for \emph{lifecycle prediction}, it remains open how a single EP method can achieve all requirements at once. It will be interesting to see, whether the current fragmentation in EP research is unavoidable due to the diverse nature of monitored systems stemming from the nine application domains (cf. \autoref{tab:domain}) and the wide range of respective research areas related to EP or can be overcome as its fragmentation is rather due to inadequate interdisciplinary research and research being too focused. To emphasize the importance of further understanding the reasons that are responsible for the fragmentation, the challenges and research directions in the next section start with the challenge of holistic EP. 

\janikrev{Overall, EP methods broadly support the requirements for building a PCM system concerning the input and output of the prediction method as well as its evaluation and understanding (cf. Fig. \ref{fig:req}). Looking more closely at the requirements support, the following observations can be made:}\\
\janikrev{
\OB{1} EP supports a selection of requirements through a combination of the prediction goal and EP method. \\
\OB{2} The number of approaches for EP has been and is steadily increasing. \\
\OB{3} None of the approaches supports all requirements; most of the approaches support one to three requirements and at most 8 out of 15 requirements. \\
\OB{4} Approaches supporting requirements with $\lambda \leq 0.04$, $0.04 < \lambda < 0.2$ and $\lambda > 0.2$ can be distinguished with distribution patterns quite clearly, i.e., selective occurrence, plateau, and steady increase. An interpretation here is that requirements that are already addressed, remain being addressed with increasing interest. Requirements that have not been addressed or only in a limited way remain being addressed with low intensity. \\
\OB{5} None of the approaches supports event life cycle prediction. \\
\OB{6} Although the support of root cause analysis (\emph{RCA support}) is fully met by the highest number of papers in the article selection, other root cause related requirements are only supported by very few approaches, i.e., \emph{RCA visualization}, \emph{precise RCA}, and \emph{joint RC}. Hence it seems that the importance of root cause analysis is acknowledged and addressed, but advanced root cause related requirements remain still open. \\
\OB{7} Although the existence of multiple event sources is supported by a quite high number of approaches, other data-related requirements such as \emph{matching} and \emph{data quality} are only addressed by very few approaches. Hence, it seems that the existence of multiple and possibly heterogeneous data sources is acknowledged and addressed, but further aspects connected to the quality of data remain still open. \\
\OB{8} Predicted events without an occurrence probability (\emph{probabilistic prediction}) and confidence (\emph{uncertainty quantification}) can give a false sense of urgency to act on low probability or confidence events. It remains to be open how the various EP methods can be advanced to account for probabilities and confidences. \\
\OB{9} Despite the importance of visualizing predicted events and root causes to the user, \emph{output visualization} and \emph{RCA visualization} are only (partly) met by roughly 80 articles and 30 articles respectively. As EP is formulated through a heterogeneous range of conceptualizations (cf. \janikrr{taxonomy in Figure 2 of the supp. material\supplementary}), it is not straightforward to deduce visualizations from the article's EP method and evaluation. \\
\OB{10} Suitable evaluation metrics for an EP method are a condition for understanding the method's qualities. Nevertheless, \emph{probabilistic quality} is only successfully addressed by less than 5\% of the articles.}

%Conclusions 1 -- 5 underpin the importance of the PCM system concept for holistic EP and point to open challenges and research directions, which will be discussed in Section \ref{sec:challenges}.
\janikrev{Overall, \OB{1} to \OB{10} point to an ongoing research effort in EP with a focus on selected requirements which results in a rather static and fragmented research landscape. Hence, the integrative effort to combine EP methods from different areas and application domains as well as taking a system perspective offers the opportunity to consider EP under a broader set of requirements. Observations \OB{1} and \OB{2} are understood as confirming our EP definition (cf. \autoref{sec:req}) and the importance of EPs. Based on \OB{3} to \OB{10}, we formulate challenges and research directions in \autoref{sec:challenges}. }

\section{Open Challenges and Research Directions}
\label{sec:challenges}
\janikrev{In order to counteract the current fragmentation in EP research, we derive open challenges from the assessment in \autoref{sec:assessment}, in particular observations \OB{3} to \OB{10}.}
%(cf. \autoref{ssec:summary}).
%To bring the challenges inline with existing surveys, we relate them with challenges from the yardstick EP survey \cite{zhao_event_2021} and the yardstick PPM survey \cite{rinderle-ma_predictive_2022} in \autoref{ssec:existingchallenges}.  } 

%\subsection{Assessment Challenges}
%\label{ssec:ourchallenges}

\noindent
\textbf{C1 - Holistic Event Prediction Method}\\
 %Despite valuable advancements in all key requirements, 
\noindent\textsl{Challenge:}
\janikrev{A direct consequence of \OB{3} is that} no EP method simultaneously meets all requirements. Hence, selecting a single EP method leaves us with the necessity of trading off various sets of \janikrev{requirements} with one another. \janikrev{In particular, some requirements like \emph{joint RCs} remain a blind spot for EP research, while other requirements like \emph{suffix prediction} are still underrepresented (cf. \OB{4}). The challenge is a development of future holistic EP methods that support more or all requirements and, thus, represent the manifold nature of EP.}
\janikrev{Moreover, the value of EP is only determined by accuracy of predicted events, as indicated by the prominent focus of EP research on accuracy. However, many more factors determine the value of EP in the PCM system as the assessment and related surveys emphasize \cite{zhao_event_2021,rinderle-ma_predictive_2022}. Aside from accuracy, the event resolution (e.g., hospital vs. ward utilization), likelihood (i.e., \emph{probabilistic prediction} in \autoref{ssec:functional}), confidence (i.e., event uncertainty), prediction timeliness, event intensity (e.g., how bad is the occurrence), explainable events (e.g., root causes can be determined) and further quality dimensions that result from the requirements in \autoref{sec:arch} are important. Additionally, to what degree the event is actionable can play a significant role, as healthcare professionals do not act, if a serious condition is predicted that require aggressive interventions \cite{eini-porat_tell_2022}.} \\
%to remove this choice and follow a holistic approach in developing further
%develop EP methods \janik{that support more or all requirements} \janikrev{through single methods or combining multiple methods}.  
\noindent\textsl{Research direction:} \janikrev{The challenge of developing holistic EP methods requires \janikrr{\textbf{interdisciplinary research}}. For instance, explainable events in the domain of manufacturing can be defined trough a collaboration of EP researchers and mechanical engineers. By gradually increasing the number of researchers and application domains in the collaboration, we may be able to push the definition of explainable events to a generic formulation.} %by adopting principles of user-centred design \cite{gulliksen2003key}, future research can explore means of presenting and visualizing the complex output of EP and the PCM system to the user (cf. \C{4}). This challenge should not be taken lightly, as it significantly determines the adoption of the method by users. 
\janikrr{
%Given a set of clearly defined quality dimensions, EP must start to acknowledge the 
EP is a \textbf{multi-objective} endeavor
%. Yet even more, EP 
and must consider the abundance of regulatory requirements on monitored systems such as banks \cite{gai_regulatory_2019}, clinics \cite{center_for_regulatory_stategy_americas_2020_2020,braithwaite_can_2017} or manufacturing facilities \cite{herrera_vidal_complexity_2021} to define \textbf{appropriately complex and realistic prediction goals}. A promising future research avenue is to deduce a \emph{markov decision process} (MDP) \cite{alagoz_markov_2010} from the prediction output. By meeting all requirements for \textbf{optimal decision-making} (cf. \autoref{sec:arch}), the prediction output is rich enough for constructing a MDP that models the future and its corresponding decisions in a statistical process. Then, the multi-objective optimization problem of EP together with formalized constraints from regulatory constraints can be combined into a goal for optimal \emph{policy synthesis} \cite{lavaei_automated_2022}. As a result, EP can benefit from the numerous solution techniques to the \textbf{optimal policy synthesis} problem, which asks for a sequence of decisions that are most likely to guarantee the specified goal given the MDP.}  %general applicability in real-world PCM systems would advance the state-of-the-art.} 
 %\janikrev{For example, by combining solution techniques for the requirements, future research on EP has sources of inspiration for advancing EP through integrating existing solution techniques.} 
 
\janikrr{In general,} for a \textbf{holistic approach to method development}, future research can either combine existing properties from various methods into a single method, develop new solutions to the existing problems for a single method, or combine existing methods and, thus, their properties in a generic framework.
Independent of the chosen option, a conceptual model of EP that identifies the systematic relationships between application domains, domain knowledge, prediction goals, input data, feature engineering, prediction methods and the action space is desirable. 
%This conceptual model particularly represents our understanding on how an application domain, the existing domain knowledge and given prediction goals require a certain set of data sources, a certain set of features engineered on the data, appropriate prediction methods and the possible action space.
Such a model extends the structural system perspective on EP and its surrounding components embodied by the PCM system (cf. \autoref{fig:framework}) and the methodological perspective on EP in \cite{zhao_event_2021} with a \textbf{behavioral and operational system} perspective.

\noindent\textbf{C2 - Semantics, Knowledge and Explainability}\\
%\label{ssec:semantic}
%\textsl{Classification:} \OB{3}, \OB{4}, \EC{1}, \EC{2}, \PC{1}, \PC{2}
 \noindent\textsl{Challenge:}
 EP aims at anticipating the future of the monitored system by means of algorithms. Monitored systems come from a diverse range of application domains and, thus, have a diverse range of semantics. At some point in the PCM system, the algorithms used to develop EP should take the semantics, e.g. of the event lifecycle (cf. \OB{5}), into account. The same applies to existing domain knowledge. Aside from potentially positive effects on accuracy, the way semantics and knowledge are taken into account determines the explainability of the method and its generality: If the method is very abstract and generic, it is typically not explainable to the domain expert, whereas a very concrete and domain-specific method is typically explainable, but does not generalize. The challenge is to shift the trade-off between explainability and generality and develop new ways to integrate semantics and knowledge in the method for an improved explainability while maintaining generality (cf. \OB{6}).\\
 \noindent\textsl{Research direction:} Starting with a clear definition of explainability for EP, e.g. is it rather a counterfactual or the more strict causal definition \cite{rice_leveraging_2021}, and the necessary content type, communication means and target group of the explanation \cite{burkart_survey_2021}, future research can evaluate the correctness of explanations. Explanation-guided learning \cite{gao_going_2022} demonstrates promising approaches to explainability and correctness evaluation, yet it pre-determines the way of how the method integrates semantics and domain knowledge by stating them as explanation supervision and regularization in the learning objective. Other or additional ways of integration could be parametrization, feature engineering or a certain hypothesis space for explainable models. It is highly beneficial to explore various combinations of integration and test for generality of the method such that we can improve our understanding on how semantics and domain knowledge can be used for explainability while maintaining generality. 
 
 \janikrev{Since the quality of predicted events is also affected by the degree to which existing mechanistic knowledge \cite{zhao_event_2021} is integrated within the prediction model, \textbf{hybrid models} are beneficial. \janikrr{Hybrid models combine expressive, even \textbf{executable models} that represent the known causal relationships on a given monitored system with the generally well-performing machine learning models \cite{greasley_enhancing_2021,DBLP:books/sp/Rinderle-MaMR24}. The result benefits from superior prediction performance and maintains transparency.} In particular, hybrid models are preferred that do not simply apply ensemble learning-based techniques, but integrate the knowledge and rules on a more fundamental level into the prediction model similar to physics-informed neural networks \cite{cuomo2022scientific}. \janikrr{Here, EP research can catch up from the related research stream that combines operations research with machine learning \cite{greasley_enhancing_2021}.} Yet, the count of EP methods with hybrid models is decreasing (cf. \autoref{ssec:aggregate}). Nevertheless, a general framework that represents a solution of how to integrate mechanistic knowledge across domains would be very beneficial. %In general, designing advanced objectives and regularization terms to account for the multi-dimensional nature of quality improves the value of EP. 
 Since it is likely that there exist trade-offs between the quality dimensions, user studies on typical trade-offs in domains or in general help to integrate real-world needs to future research.}

\noindent\textbf{C3 - Flexibility and Robustness}\\
%\label{ssec:flex}
%\noindent\textsl{Classification:} \OB{5}, \OB{4}, \EC{1}, \PC{1}
\noindent\textsl{Challenge:}
Considering the heterogeneity of application domains, data sources and prediction goals, EP methods must find a balance between flexibility and robustness \janikrev{(cf. \OB{7} and \OB{8})}. On the one hand, they have to be flexible with respect to \janikrev{three aspects}. \janikrev{First,} domain-specific properties and requirements (i) \janikrev{captures the effects of the respective application domain and environment}. \janikrr{For example, compare} a very dynamic, clinical environment in the healthcare domain with a particularly strong requirement on uncertainty vs. a relatively stable environment of a standardized goods manufacturing facility with a relatively weak requirement on uncertainty. \janikrev{Second,} the data properties and quality issues of multiple data sources (ii) \janikrev{captures the effects of the available data sources}, e.g. different granularities of the data such as city- vs. country-level and missing data. \janikrev{Third,} prediction goal properties (iii) \janikrev{captures the effects of how the future should look like}, e.g. a single, stable prediction goal such as "treatment outcome of a patient is positive" vs. multiple, dynamic prediction goals such as the set of regulatory constraints a bank's business processes have to adhere to. 
In addition to flexibility, EP methods have to be robust with respect to adversarial attacks and/or noisy input data \cite{comuzzi_does_2019,klinkmuller_towards_2018,zhao_event_2021}, since EP in a PCM system has a direct real-world consequence through the action that the user chooses. \\
\noindent\textsl{Research direction:} One way to deal with the flexibility challenge is to limit the scope of the method to a certain problem space for which flexibility is not required anymore. So far, this approach dominates research on EP. Although this approach can solve the challenge by removing it from the equation to a large extent, it results in a very fragmented set of EP methods and transfers the challenge to the user. Future methods should abstain from definitions that simplify the problem in this way. For (i) domain-specific properties and requirements and (ii) data properties and quality issues existing work supports the required flexibility (cf. assessment for \emph{probabilistic prediction}, \emph{uncertainty quantification}, \emph{online conceptualization}, \emph{matching}, \emph{multiple sources} in \autoref{sec:assessment}). This combination is promising for future work on meeting flexibility for (i) and (ii) together. While \cite{DBLP:journals/is/LyMMRA15} called for flexibility regarding various sets of prediction goals, flexibility for (iii) prediction goal properties is usually still abstracted from or left unmentioned. Considering the abundance of regulatory requirements on monitored systems (cf. \textbf{C1}), EP methods that are flexible with respect to prediction goal heterogeneity are highly beneficial. \janikrr{Once again, \textbf{hybrid models} present themselves to separate predicting future events from the subsequent matching of prediction goals on the predicted sequence of future events \cite{rinderle-ma_predictive_2022}. As the hybrid model does not directly predict the prediction goal, it is significantly more flexible towards changing goals. }
Yet, a more flexible EP leaves more room for adversarial attacks. Consequently, research on flexible EP should study how these mechanisms interact with adversarial attack models. Considering noise, methods are desirable that can reliably distinguish variances in the input data that require flexibility from noise that \janik{should be captured as uncertainty}.

\noindent\textbf{C4 - User Interface}\\
%\label{ssec:user}
\noindent\textsl{Challenge:}
As EP research focuses on the method and its rather technical evaluation, it often does not provide a simple and clear user interface of relevant output (cf. \OB{9}). Moreover, probabilistic predictions with uncertainty estimates, multiple prediction goals, root causes and potential actions aggravate the challenge to present the output in a simple and clear way to the user. Although the user interface corresponds to the output component of a PCM system that is separated from the prediction component, the method design should already have the subsequent user interface in mind to ease its design and implementation, e.g., by suggesting certain visualization techniques. Lastly, some methods have started proposing the integration of prediction with decision theory to include actions and their impact \cite{soleimani_scalable_2018,wang_anomaly_2019,bozorgi_prescriptive_2021}, but these are goal-dependent and domain-specific and, thus, cannot be easily transferred to other goals and application domains. Furthermore, \cite{soleimani_scalable_2018,wang_anomaly_2019,bozorgi_prescriptive_2021} lack to provide how the additional information should be integrated to present the output to the user. \\
\noindent\textsl{Research direction:} By adopting principles of \textbf{user-centered design} \cite{gulliksen2003key}, future research \janikrr{should} explore means of presenting and visualizing the complex output of EP and the PCM system to the user. This challenge should not be taken lightly, as it significantly determines the adoption of the method by users and may affect the user's trust in the system. Research that incorporates actions through a decision theoretic framework further supports the user in deciding what to do based on the anticipated future. By allowing for parametrizing the action space (cf. \textbf{C1}), future work on EP can increase the flexibility of the method (cf. \textbf{C3}) and enables the user in setting the action space. The development of a \textbf{comprehensive user interface} for a PCM system that streamlines the relevant information to the user and takes subsequent actions into account is key for building trust and increasing the accessibility for users.

\noindent\textbf{C5 - Evaluation}\\
%\label{ssec:eval}
\noindent\textsl{Challenge:}  \cite{zhao_event_2021} presents the evaluation design with respect to matching predicted with real events and metrics of effectiveness for EP, but neither considers probabilistic metrics of effectiveness and other dimensions of EP nor the performance in terms of training and prediction time. However, both are crucial for a standardized and valuable evaluation of EP. \janikrev{Without suitable evaluation metrics (cf. \OB{10}), we cannot ascertain how successful an EP method is in achieving its design goals and how it compares to its peers. Besides, optimizing the prediction model with respect to unsuitable evaluation metrics skews the method's predicted events in the wrong direction such that we cannot derive practically meaningful conclusions from the evaluation results. Additionally, it is not clear to what extent evaluation metrics are linked to actual user value and currently popular evaluation metrics fail to measure how well a visualization presents predicted events and root causes (cf. \textbf{C4}).  Hence, the challenge is to carefully design the evaluation, ideally, with users, and new metrics such that we can draw reliable conclusions for the state-of-the-art and the application.} \\
\noindent\textsl{Research direction:} \cite{doi:10.1287/deca.2016.0337} reviewed 201 probabilistic metrics of effectiveness that are candidates for EP. Both an analysis as well as standard proposal for EP is needed. \janikrev{However, probabilistic metrics only account for the probabilistic nature of EP. To also account for the other dimensions such as event resolution, \janikrr{timeliness,} confidence, intensity, explainability, and flexibility (cf. \textbf{C1} - \textbf{C3}), new metrics have to be developed that build on the standard probabilistic metrics. \janikrr{One promising direction is to formalize the desired qualities like timeliness as \textbf{real-time constraints} in a continuous-time logic \cite{lavaei_automated_2022}. Both evaluating the satisfiability given the prediction model and computing a likelihood for satisfaction should be achievable in a controlled experimental setting. }During the development of new metrics, \textbf{interdisciplinary work} is desired such that the metric design reflects perspectives from, e.g., lawyers for explainable events. The resulting new metrics} can be directly applied in a comprehensive, empirical benchmark of existing methods within comparable classes of our taxonomy for EP methods (cf. Section B in the supp. material\supplementary). An empirical benchmark that allows for \textbf{simple extension} with newly developed EP methods is beneficial to support future evaluation of methods \janikrr{similar to \cite{shao_exploring_2025}}. For evaluating visualization (cf. \textbf{C4}) and the design of a user interface, user studies have to be carried out. %With respect to the time performance, the comparison can utilize appropriate benchmark case studies to additionally compare the training and prediction times.

\section{Conclusion, Impact, and Outlook}
\label{sec:conclusion}
\janikrev{This work presents a comprehensive survey on EP methods spanning the disparate research areas of EP and PPM. Taking this umbrella review approach allows to draw novel challenges and research directions from a set of observations that we take on the result of assessing existing EP methods with an integrated set of requirements. 
%As a consequence,  that hold across research areas and application domains. 
%and to put existing ones from yardstick surveys to the test. 
Moreover, taking a system perspective lifts the survey from (technical) details of single EP methods to an integrated view in a PCM system. This system approach can be taken as blueprint to assess the realization of real-world project that aim to apply EP for decision support. %as shown by the examples of healthcare and manufacturing. 
The holistic approach with a systems perspective reveals challenges regarding the joint application of EP methods, their ability to deal with uncertainty in highly flexible environments, the evaluation of EP quality, and the inclusion of users based on advanced root cause analysis, visualization, and user interfaces. }

\janikrev{This survey creates particular impact by knowledge sharing (cf. Section F in the appendix for background). }
\janikrev{
%Our work positively affects knowledge sharing with respect to mediating factors and mechanisms. 
First, this work positively mediates mechanisms by means of a shared mental model and language. The PCM approach integrates the analysed EP methods, and the assessment pertains to a shared set of requirements. 
%This establishes a shared mental model. Through mapping different terminology to a standard set of terminology, we define a shared language. Thus, this work benefits knowledge sharing independent of specific knowledge sharing mechanisms by improving comprehension of EP across domains and research areas. 
Second, this work establishes a tool for several knowledge sharing mechanisms, both for academia and organizations. The classification and assessment is available in our online repository and can be used
%. Thus, the classification and assessment constitute a domain and research area spanning knowledge base for EP knowledge sharing. Having a knowledge base of shared materials is an influential factor in effective knowledge sharing among academic faculty \cite{seonghee_analysis_2008}. Moreover, the PCM approach, its requirements, and the illustration of EP methods in the assessment can be a source 
for, e.g., creating instructional material.} %\cite{seonghee_analysis_2008} used for teaching. }
%For organizations, this work, particularly the PCM approach constitutes a reference model in and tool for the three organizational mechanisms formal interaction (M2), informal interaction (M3), and communities of practice (M4). For example, during workshops of communities of practice, the PCM approach can be used as a visual tool as demonstrated in \autoref{sec:realizations}.}
\janikrev{
%The impact of effective knowledge sharing differs for academia and organizations. For researchers, effective knowledge sharing is related to increased productivity as a result of less time spent in teaching \cite{seonghee_analysis_2008}. Therefore, the increased productivity indirectly affects future development of EP methods by freeing time that can be allocated towards research. Furthermore, 
Moreover, the survey at hand acts as a starting point to access the vast literature on EP methods in its application domains and disparate research areas. Additionally, it gives a comprehensive, state-of-the-art overview such that further research directions can be easily derived. For practitioners, the PCM system can be used as a blueprint to analyze their specific requirements; this is illustrated for two application domains in \autoref{sec:realizations} and Section E of the supp. material.}
%effective knowledge sharing is related to increased business performance \cite{farooq_conceptual_2018}. To sum up, our work bridges the gap between disparate research areas by means of knowledge sharing. Knowledge sharing for researchers is particularly facilitated through a shared language and mental model and providing a knowledge base and tool for mechanisms.}

\janikrev{In future work, we plan for applying the PCM approach in additional application scenarios, together with practitioners, and for conducting user studies for EP in order to validate research goals with real-world value and to create extended datasets that, for example, include actions. 
%For practitioners, next steps are applying the PCM approach to your own project as demonstrated in Section 5. 
%From Section 4, practitioners can get inspiration for best practices and from Section 6, the expectations on EP are rooted in the state-of-the-art. For both researchers and practitioners, we have added the second-to-last paragraph in Section 1.2 to include pointers to supplementary material for studying. Also, have a look at our Git repository as a resource and approach us for further insights on our experience from industry projects
Moreover, we plan to undertake interdisciplinary work in order to overcome the complex nature of EP across domains and research areas such as law for explainability, design for the output component and visualization, as well as decision theory for prescriptive and counterfactual analysis. }

\begin{acks}
This work has been funded by 
the \grantsponsor{514769482}{Deutsche Forschungsgemeinschaft (DFG)}{https://www.dfg.de/} under project number \grantnum{514769482}{514769482}.
\end{acks}

\bibliographystyle{ACM-Reference-Format}
%\bibliography{bib,usecases}


\end{document}